\documentclass[acmsmall,screen]{acmart}
\usepackage{multirow, paralist, color,amsmath,amsthm}

\AtBeginDocument{%
  \providecommand\BibTeX{{%
    \normalfont B\kern-0.5em{\scshape i\kern-0.25em b}\kern-0.8em\TeX}}}

\setcopyright{acmcopyright}
\copyrightyear{2022}
\acmYear{2022}

\acmJournal{CSUR}
\acmVolume{}
\acmNumber{}
\acmArticle{}

\def\z{{\bf z}}
\def\O{\mathcal{O}}
\def\P{\mathcal{P}}

\def\S{\mathcal{S}}

\def\R{\mathbb{R}}

\def\M{\mathcal{M}}

\def\I{\mathbb{I}}

\def\EX{{\mathbb{E}}}

\def\R{\mathbb{R}}

\def\p{\mathbf{p}}

\def\x{{\bf x}}

\def\bI{{\bf I}}

\def\w{\mathbf{w}}
\def\v{\mathbf{v}}

\def\v{\mathbf{v}}

\usepackage{bm}

\def\EX{{\mathbb{E}}}

\def\R{\mathbb{R}}
\def\bI{{\bf I}}

\def\bw{\mathbf{w}}

\def\v{\mathbf{v}} 
\def\u{\mathbf{u}}

\def\EX{\mathbb{E}}

\usepackage[graphicx]{realboxes}
\usepackage{makecell}
\usepackage{algorithm,algorithmic}
\newcommand\numberthis{\addtocounter{equation}{1}\tag{\theequation}}

\begin{document}

\title{AUC Maximization in the Era of Big Data and AI: A Survey}

\author{Tianbao Yang}
\affiliation{%
  \institution{Texas A\&M University}
  \city{College Station}
  \country{USA}}
\email{tianbao-yang@uiowa.edu}

\author{Yiming Ying}
\affiliation{%
  \institution{University at Albany, SUNY}
  \city{Albany}
  \country{USA}}
  \email{yying@albany.edu}

\renewcommand{\shortauthors}{Yang and Ying}

\begin{abstract}
Area under the ROC curve, a.k.a. AUC, is a measure of choice for assessing the performance of a classifier for imbalanced data. AUC maximization refers to a learning paradigm that learns a predictive model by directly maximizing its AUC score. It  has been studied for more than two decades dating back to late 90s and a huge amount of work has been devoted to AUC maximization since then. Recently, stochastic AUC maximization for big data and deep AUC maximization (DAM) for deep learning have received increasing attention and yielded dramatic impact for solving real-world problems. However, to the best our knowledge there is no comprehensive survey of related works for AUC maximization.  This paper aims to address the gap by reviewing the literature in the past two decades. We not only give a holistic view of the literature but also present detailed explanations and comparisons of different papers from formulations to algorithms and theoretical guarantees. We also identify and discuss remaining and emerging issues for DAM, and provide suggestions on topics  for future work. 
\end{abstract}

\ccsdesc[500]{Computing methodologies~Machine learning algorithms}
\ccsdesc[500]{Theory of computation~Continuous optimization}
\ccsdesc[500]{Theory of computation~Stochastic control and optimization}
\keywords{AUC, ROC, big data, deep learning}

\maketitle

\section{Introduction}
 ROC (receiver operating characteristic)  curve is a curve of true positive rate (TPR, equivalently sensitivity or recall)  versus false positive rate (FPR, equivalently fall-out)  of a classifier by varying the threshold.    The method was originally developed for operators of military radar receivers starting in 1941~\cite{GreenSwets66}.  ROC analysis has emerged as an important tool in many domains, e.g., medicine, radiology, biometrics, meteorology, forecasting of natural hazards, and is widely used in machine learning and artificial intelligence. A statistic measure associated with the ROC curve if the area under the curve (AUC), which has been widely used for assessing the performance of a classifier. Another closely related measure is called partial AUC, which refers to AUC in a certain region that restricts the range of FPR and/or TPR. 

 A standard approach in machine learning for learning a predictive model is to optimize some performance metric. A traditional performance metric of a classifier is the accuracy, i.e., the proportion of examples that are predicted correctly. However, accuracy can be misleading when the data is imbalanced, meaning that the number of data points from one class is much larger than the number of data points from the another class. In contrast, AUC is a more informative measure than accuracy for imbalanced data. However, studies show that algorithms that maximize accuracy of a model does not necessarily maximize the AUC score~\cite{Cortes2003AUCOV}. Hence, it is necessary to study algorithms for maximizing AUC directly. 
 
 
 AUC maximization in machine learning has a long history dating back to late 90s~\cite{Herbrich1999d}. Tremendous studies have been devoted to this topic and various aspects have been studied ranging from formulations to algorithms and theories. Below, we give a brief overview with exemplar references.   First, AUC maximization has been studied in the context of different learning paradigms, e.g., supervised learning~\cite{Joachims,Steck2007HingeRL},  semi-supervised learning~\cite{shijunsemiAUC,Iwata2020SemiSupervisedLF}, positive-unlabeled (PU) learning~\cite{DBLP:journals/ml/SakaiNS18,DBLP:journals/corr/abs-1803-06604}, active learning~\cite{4053043,10.1016/j.neucom.2010.01.001}, Bayesian learning~\cite{DBLP:conf/ecai/Gonen16}, federated learning~\cite{DBLP:conf/icml/GuoLYSLY20,DBLP:conf/icml/YuanGXYY21}, online learning~\cite{Zhaoicml11,gao2013one}.  Second, models in different forms have been learned in the context of AUC maximization, including linear models~\cite{ying2016stochastic}, kernel models~\cite{Herbrich1999d,Pahikkala2008EfficientAM}, extreme learning machines~\cite{YANG201774}, decision trees~\cite{10.5555/945365.964285}, neural networks~\cite{10.5555/3041838.3041945}, deep neural nets~\cite{yuan2022compositional}. Third, various solvers based on different methodologies have been studied, e.g., linear programming~\cite{nortonbuffauc}, quadratic programming~\cite{Herbrich1999d}, cutting-plane methods~\cite{Joachims}, L-BFGS~\cite{ledell16},  evolutionary algorithms~\cite{10.1007/978-3-642-13769-3_41}, gradient descent methods~\cite{herschtal2004optimising}, stochastic gradient methods~\cite{ying2016stochastic}, other methods~\cite{Bostrm2004PruningAE,calders2007efficient}. Fourth, different theoretical guarantees have been examined, e.g., consistency~\cite{10.5555/2832249.2832379}, generalization error bounds~\cite{lei2020sharper}, excess risk bounds~\cite{guo2017online,ying2016online}, regret bounds~\cite{Zhaoicml11}, convergence rates or sample complexities~\cite{liu2019stochastic}, stability~\cite{lei2021generalization,yang2021simple}. Last but not least,  AUC maximization has been successfully investigated in a variety of  applications~\cite{8731461,10.1145/3411408.3411422,DBLP:journals/ijitdm/ZhouLY09,DBLP:journals/bigdata/YamaguchiMMU20,10.1093/bioinformatics/btw446,10.2312/vcbm.20171246,10.1093/bioinformatics/btx255,kyumbm13,DBLP:journals/corr/abs-1202-3701,Feizi2020HierarchicalDO,10.5555/2887007.2887048,DBLP:conf/pkdd/WangSX16}, e.g., medical image classification~\cite{DBLP:journals/corr/abs-2012-03173} and molecular properties prediction~\cite{wang2020moleculekit},  to mention but a  few. 
 
\begin{figure}[t]
\begin{center}
\centerline{\includegraphics[width=0.9\columnwidth]{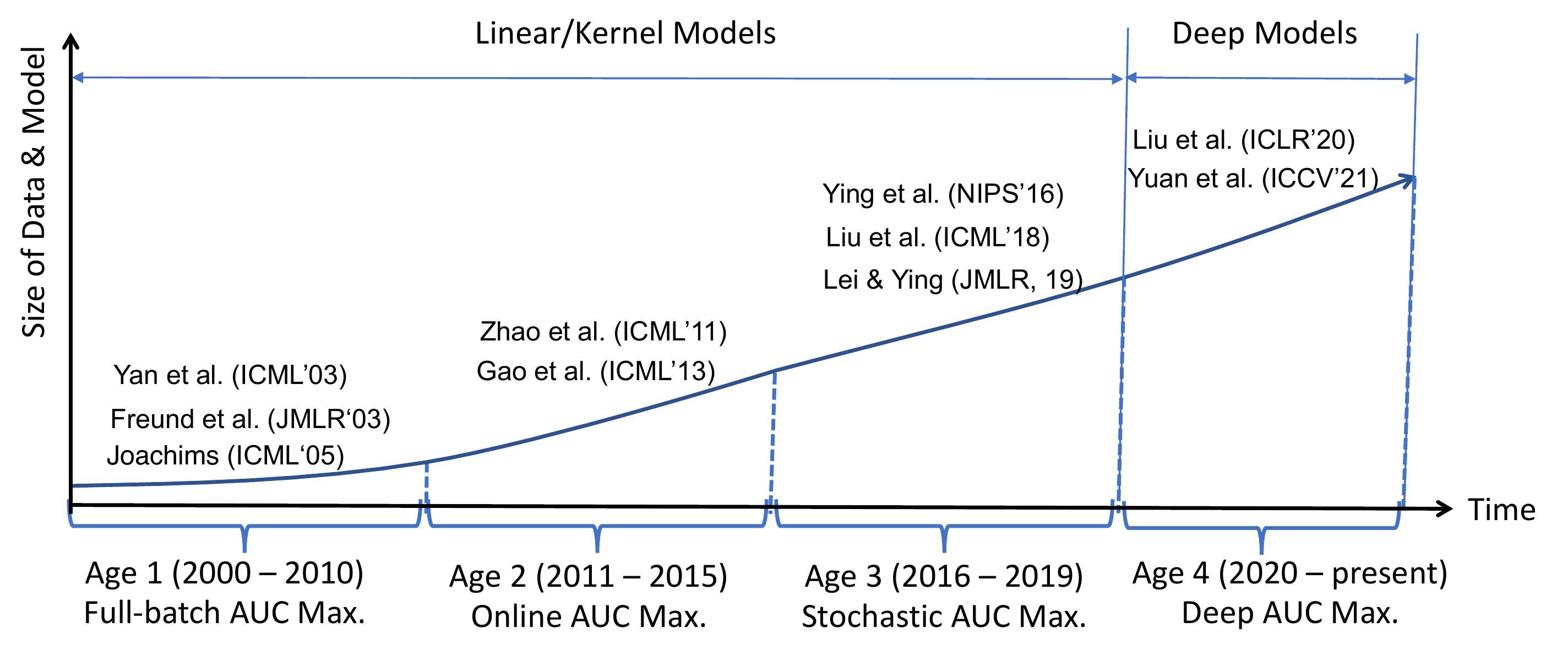}}
\vspace{-0.14in}
\caption{Four ages of AUC maximization and exemplar works in each stage.}
\label{fig:hist}
\vspace*{-0.1in}
\end{center}
\end{figure}

A bulk of studies related to AUC maximization revolve around the development of the solver, i.e., optimization algorithms, for learning a predictive model. The reason is that compared with the traditional metric of accuracy, the AUC score is non-decomposable over individual examples, which renders its optimization much more challenging, especially for big data.  The research of AUC maximization algorithms has experienced four different ages in the long history of  two decades, namely full-batch based methods for the first age (roughly 2000 - 2010), online methods for the second age (roughly 2011 - 2015), stochastic methods for the third age (roughly 2016 - 2019), and deep learning methods for the recent age (roughly 2020 - present). The first three ages focus on learning linear models or kernelized models, and the last age focuses on deep neural networks.  In each age, there have been seminal works in rigorous optimization algorithms that play important roles in the evolution of AUC maximization methods.  The four ages are illustrated in Figure~\ref{fig:hist}.

To the best of our knowledge, there is no comprehensive survey devoted to AUC maximization. The only related survey work  is~\cite{surveyor11} published in 2011. Nevertheless, it focuses on ordinal regression and does not provide a comprehensive survey of optimization algorithms for AUC maximization with theoretical guarantees.  This paper aims to address this gap by providing a comprehensive review of related works for AUC maximization, with a particular focus on the optimization algorithms. We will cover important works in all four ages about the optimization algorithms and discuss their properties.  The remainder of this paper is organized as follows. 
\begin{itemize}
    \item We provide some background for AUC and AUC estimators in Section~\ref{sec:bac}. We give definitions for both AUC and partial AUC and derive their non-parametric estimators. 
    \item In Section~\ref{sec:obj}, we review different objective functions for AUC maximization, and mainly discuss three families of objectives. 
    \item We review full-batch based methods for solving AUC maximization in the first age for both AUC maximization and partial AUC maximization in Section~\ref{sec:full}. 
    \item In Section~\ref{sec:onl}, we present two classes  of online optimization methods for AUC maximization and discuss their properties. 
    \item We present stochastic optimization methods in both offline setting and online setting in Section~\ref{sec:sto}, and compare their properties. 
    \item In Section~\ref{sec:dam}, we survey recent papers about non-convex optimization for deep AUC and partial AUC maximization, and discuss their applications in the real world. 
    \item In Section~\ref{sec:out} we discuss remaining and emerging issues in deep AUC maximization, and provide suggestions of topics for future work. Finally, we conclude in Section~\ref{sec:con}.
\end{itemize}

{\bf Disclaimer.} Before ending this section we would like to point out that we have done our best to include as many related works in machine learning as possible, and may innocently miss some relevant papers in machine learning or other areas. We also emphasize that this paper is about maximization of areas under ROC curves and does not cover the maximization of areas under Precision-Recall curves. Finally, we present a list of three fundamental papers of AUC, top 10 Cited Papers (as of 07/28/2022)  related to AUC maximization, and two representative works for deep AUC maximization in Table~\ref{tab:fund}.

\begin{table*}[t] 
	\caption{Three fundamental papers of AUC, top 10 Cited Papers related to AUC maximization and two representative works for deep AUC maximization. 
	The citations data is from Google Scholar as of July 28, 2022, which is included for reference and is by no means the only metric to measure the influence of a paper.  {\bf Remark}: we do not include the highly cited paper~\cite{10.5555/945365.964285} in the list due to that it was studied in~\cite{Cortes2003AUCOV} for AUC maximization.   }\label{tab:fund} \vspace*{-0.1in}
	\centering
	\label{tab:loss}
	\scalebox{0.58}{\begin{tabular}{lllccc}
			\toprule
	        Title&Authors&Year & Citations&Venue&Reference\\
		    \hline 
		    \makecell[l]{1. The meaning and use of the area under a receiver\\ operating characteristic}&J. A. Hanley and B. J. McNeil.&1982 & 22314&Radiology&\cite{Hanley1982}\\
			2. Analyzing a Portion of the ROC Curve&D. Katzman McClish&1989&769&
Med Decis Making&\cite{doi:10.1177/0272989X8900900307}\\
			 3. Partial AUC Estimation and Regression& L. Dodd and M. Pepe&2003&389&Biometrics&\cite{dodd03}\\
			\midrule
			\midrule
		    1. A support vector method for multivariate performance measures& T. Joachims&2005&1022&ICML&\cite{Joachims}\\
		    2. AUC Optimization vs. Error Rate Minimization&C. Cortes and M. Mohri&2003&721&NIPS&\cite{Cortes2003AUCOV}\\
		     \makecell[l]{3. Optimizing Classifier Performance via an
Approximation\\ to the Wilcoxon-Mann-Whitney Statistic}& \makecell[l]{L. Yan, R. Dodier, \\M. C. Mozer, and R. Wolniewicz}& 2003& 350&ICML&\cite{10.5555/3041838.3041945}\\
		    4. Optimising area under the ROC curve using gradient descent&A. Herschtal and B. Raskutti&2004&231&ICML&\cite{herschtal2004optimising}\\
            5. Online AUC Maximization &\makecell[l]{P. Zhao, S. C. H. Hoi, R. Jin, and T. Yang}& 2011&  217&ICML&\cite{Zhaoicml11}\\
		    6. AUC maximizing support vector learning&U. Brefeld, T. Scheffer& 2005& 184&ICML (workshop)&\cite{brefeld2005auc}\\
		     \makecell[l]{7. The P-Norm Push: A Simple Convex Ranking Algorithm\\ that Concentrates at the Top of the
List}& C. Rudin& 2009 & 179&JMLR&\cite{JMLR:v10:rudin09b}\\
		      8. One-pass AUC optimization&  W. Gao, R. Jin, S. Zhu, and Z. Zhou& 2013& 153&ICML&\cite{gao2013one}\\
9. Efficient AUC optimization for classification&T. Calders and S. Jaroszewicz &2007&  128&PKDD&\cite{calders2007efficient}\\
    10. Stochastic online AUC maximization&Y. Ying, L. Wen, and S. Lyu&2016&104&NIPS&\cite{ying2016stochastic}\\
    \midrule
    1. Stochastic AUC Maximization with Deep Neural Networks&M. Liu, Z. Yuan, Y. Ying, and T. Yang. &2020&35&ICLR&\cite{liu2019stochastic}\\
    \makecell[l]{2. Robust Deep AUC Maximization: A New Surrogate Loss and Empirical\\ Studies on Medical Image Classification}&Z. Yuan, Y. Yan, M. Sonka and T. Yang. &2021&13&ICCV&\cite{DBLP:journals/corr/abs-2012-03173}\\
\bottomrule
	\end{tabular}}
	\vspace*{-0.15in} 
\end{table*}

\section{Background}~\label{sec:bac}
{\bf Notations.} Let $\mathbb I(\cdot)$ be an indicator function of a predicate, and $[n]=\{1,\ldots, n\}$.  Let $\z = (\x, y)$ denote an input-output pair, where $\x\in\mathcal X $ denotes the input data and $y\in\{1,-1\}$ denotes its class label. Let $\P_+$ denote the distribution of positive examples and $\P_-$ denote the distribution of negative examples. Let $f(\x):\mathcal X\rightarrow \R$ denote a predictive function to be learned. It is usually parameterized by a vector $\w\in\R^d$ and we use the notation $f_\w(\cdot)$ to emphasize that it is a parameterized model. Let $\ell(\w; \x, \x') = \ell(f_\w(\x') - f_\w(\x))$ denote a pairwise loss for a positive-negative pair $(\x, \x')$.

For a set of given training examples $\S=\{(\x_i, y_i), i\in[n]\}$ in the offline setting, let $\S_+$ and $\S_-$ be the subsets of $\S$ with only positive  and negative examples, respectively, and let $n_+=|\S_+|$ and $n_-=|\S_-|$ be the number of positive and negative examples, respectively. Denote by  $\S^{\downarrow}[k_1,k_2]\subseteq\S$ the subset of examples whose rank in terms of their prediction scores in the descending order are in the range of $[k_1, k_2]$, where $k_1\leq k_2$. Similarly, let $\S^{\uparrow}[k_1,k_2]\subseteq\S$ denote the subset of examples whose rank in terms of their prediction scores in the ascending order are in the range of $[k_1, k_2]$, where $k_1\leq k_2$. 
We denote by $\EX_{\x\sim\S}$  the average over $\x\in\S$. Let $D(\p, \mathbf q)$ denote the KL divergence between two probability vectors. Let $\Delta$ denote a simplex of a proper dimension, and $\Pi_{\Omega}[\cdot]$ denote the standard Euclidean projection onto a set $\Omega$. 

\subsection{Definitions of AUC, partial AUC, two-way partial AUC} In this subsection, we fix the predictive model $f(\cdot)$ and present the definitions and formulas for computing AUC of $f$.    For a given threshold $t$,  the true positive rate (TPR) can be written as $\text{TPR}(t) = \Pr(f(\x)>t|y = 1) = \EX_{\x\sim \P_+}[\I(f(\x)>t)]$, and the false positive rate (FPR) can be written as $\text{FPR}(t) = \Pr(f(\x)>t | y=-1) = \EX_{\x\sim\P_-}[\I(f(\x)>t)]$. Let $F_-(t) = 1 - \text{FPR}(t)$ denote the cumulative density function of the random variable $f(\x)$ for $\x\sim\P_-$.  Let $p_-(t)$ denotes its corresponding probability density function. Similarly, let $F_+(t) = 1- \text{TPR}(t)$ and $p_+(t)$ denote the cumulative density function and the probability density function of $f(\x)$ for $\x\sim \P_+$, respectively. 

For a given $u\in[0,1]$, let $\text{FPR}^{-1}(u) = \inf\{t\in\R: \text{FPR}(t)\leq u\}$.  The ROC curve is defined as $\{u, \text{ROC}(u)\}$, where $u\in[0,1]$ and $\text{ROC}(u) = \text{TPR}(\text{FPR}^{-1}(u))$. The AUC score of $f$ is given by 
\begin{align}\label{eqn:aucd1}
    \text{AUC}(f)&  = \int^{1}_0\text{ROC}(u) du  = \int^{\infty}_{-\infty}\text{TPR}(t)dF_-(t)= \int^{\infty}_{-\infty}\text{TPR}(t)p_-(t)dt\notag\\
    &  = \int^{\infty}_{-\infty}\int^{\infty}_t p_+(s)ds p_-(t)dt  = \int^{\infty}_{-\infty}\int^{\infty}_{-\infty} p_+(s) p_-(t)\I (s> t)ds dt.
\end{align}
The above expression also gives a probabilistic interpretation of AUC~\cite{Hanley1982}, i.e., 
\begin{align}\label{eqn:aucd2}
\text{AUC}(f) = \Pr(f(\x_+)> f(\x_-)) = \EX_{\x_+\sim\P_+, \x_-\sim\P_-}[\I(f(\x_+)>f(\x_-))]. 
\end{align}
\begin{figure}[t]
\begin{center}
\centerline{\includegraphics[width=0.9\columnwidth]{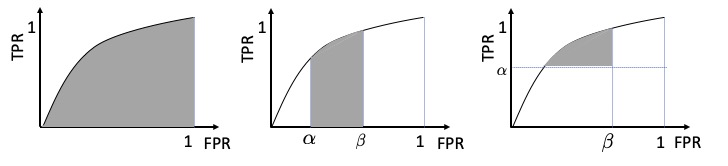}}
\vspace{-0.17in}
\caption{From left to right: AUC,  one-way pAUC, two-way pAUC}
\label{fig:auc}
\vspace{-0.15in}
\end{center}
\end{figure}
The normal AUC measure could be misleading when the data is highly imbalanced. In many applications (e.g., medical diagnostics), we would like to control the FPR in a certain range, e.g., $\text{FPR}\in (\alpha, \beta)$. Hence, another measure of interest is  {\bf partial AUC (pAUC)} with FRP restricted in the range $(\alpha, \beta)$, which is given by 
\begin{align}\label{eqn:paucd1}
    \text{pAUC}(f, \alpha, \beta)&  
    = \int^{\text{FPR}^{-1}(\alpha)}_{\text{FPR}^{-1}(\beta)}\text{TPR}(t)dF_-(t)
    = \int^{\text{FPR}^{-1}(\alpha)}_{\text{FPR}^{-1}(\beta)}\int^{\infty}_{-\infty} p_+(s) p_-(t)\I (s> t)ds dt.
\end{align}
This  expression gives a probabilistic interpretation of pAUC, which was first shown in~\cite{dodd03}, i.e., 
\begin{align}\label{eqn:paucd2}
\text{pAUC}(f, \alpha, \beta) =\Pr(f(\x_+)> f(\x_-), f(\x_-)\in[\text{FPR}^{-1}(\beta), \text{FPR}^{-1}(\alpha)]). 
\end{align}
In contrast to pAUC defined above that is also referred to as one-way pAUC, two-way pAUC has been also studied~\cite{doi:10.1177/0962280217718866}. A two-way pAUC is defined by specifying an upper bound $\beta$ on the FPR and a lower bound on $\alpha$ on the TPR. Then, the two-way pAUC (TPAUC) is given by 
\begin{align}\label{eqn:tpaucd}
\text{TPAUC}(f, \alpha, \beta) = \Pr(f(\x_+)> f(\x_-), f(\x_-)\geq  \text{FPR}^{-1}(\beta), f(\x_+)\leq \text{TPR}^{-1}(\alpha)\}). 
\end{align}
An illustration of AUC, one-way partial AUC and two-way partial AUC is given in Figure~\ref{fig:auc}. 

\subsection{Non-Parametric Estimators} Given a set of examples $\S=\S_+\cup\S_-$, how can we estimate AUC and pAUC? There are parametric estimators assuming the prediction scores following a particular distribution (e.g., normal distribution)~\cite{doi:10.1177/0272989X8900900307}, and non-parametric estimators that do not make any assumptions regarding the distribution of prediction scores. We will focus on non-parametric estimators below since they are widely used for AUC maximization. 

According to the probabilistic interpretation of AUC in~(\ref{eqn:aucd2}), a non-parametric estimator can be computed as follows that corresponds to the  Mann-Whitney U-statistic~\cite{Hanley1982}:  
\begin{align}\label{eqn:eaucd}
{\text{AUC}}(f; \S) = \frac{1}{n_+}\frac{1}{n_-}\sum_{\x_i\in\S_+}\sum_{\x_j\in\S_-}\I(f(\x_i)>f(\x_j)). 
\end{align}
A (non-normalized) non-parametric estimator of pAUC can be computed by~\cite{dodd03}:
\begin{align*}
{\text{pAUC}}(f,\alpha, \beta; \S) = \frac{1}{n_+}\frac{1}{n_-}\sum_{\x_i\in\S_+}\sum_{\x_j\in\S_-}\I(f(\x_i)>f(\x_j), f(\x_j)\in(q_{\beta}, q_{\alpha})). \end{align*}
where $q_\alpha$ denotes the $\alpha$ quantile of $f(\x_-), \x_-\sim \P_-$. The quantiles $q_{\alpha}, q_{\beta}$ are usually replaced by their empirical estimations, which gives the following non-normalized estimator of pAUC: 
\begin{align}\label{eqn:epaucd2}
{\text{pAUC}}(f, \alpha, \beta; \S) = \frac{1}{n_+}\frac{1}{n_-}\sum_{\x_i\in\S_+}\sum_{\x_j\in\S^\downarrow_-[k_1+1, k_2]}\I(f(\x_i)>f(\x_j)). \end{align}
where $k_1=\lceil n_-\alpha\rceil, k_2 = \lfloor n_-\beta \rfloor$. Similarly, a (non-normalized) non-parametric estimator for two-way pAUC is given by 
\begin{align}\label{eqn:etpaucd2}
{\text{TPAUC}}(f, \alpha, \beta; \S) = \frac{1}{n_+}\frac{1}{n_-}\sum_{\x_i\in\S_+^{\uparrow}[1, k_1]}\sum_{\x_j\in\S^\downarrow_-[1, k_2]}\I(f(\x_i)>f(\x_j))
\end{align}
where  $k_1=\lfloor n_+\alpha\rfloor, k_2 = \lfloor n_-\beta \rfloor$.

\section{Surrogate Objectives for AUC Maximization} \label{sec:obj}
\vspace*{-0.02in}\paragraph{\bf Objectives based on a Pairwise Surrogate Loss.} For AUC maximization, one often replaces  the indicator function in the non-parametric estimators of AUCs defined above by a surrogate loss function $\ell(f_\w(\x_-) - f_\w(\x_+))$ of $\I(f_\w(\x_-)\geq f_\w(\x_+))$ to formulate the objective function. As a result, AUC maximization can be formulated as  
\begin{equation}\label{eq:auc-emp}
\min_{\bw\in \R^d} \frac{1}{n_+n_-} \sum_{\x_i\in \S_+}\sum_{\x_j\in \S_-}\ell(f_\bw (\x_j) - f_\bw(\x_i)),
\end{equation} 
and (one-way) pAUC maximization can be formulated as 
\begin{equation}\label{eq:pauc-emp}
\min_{\bw\in \R^d} \frac{1}{n_+}\frac{1}{n_-}\sum_{\x_i\in\S_+}\sum_{\x_j\in\S^\downarrow_-[k_1+1, k_2]}\ell(f_\bw (\x_j) - f_\bw(\x_i)),
\end{equation} 
and two-way pAUC maximization can be formulated as
\begin{equation}\label{eq:tpauc-emp}
\min_{\bw\in \R^d} \frac{1}{n_+}\frac{1}{n_-}\sum_{\x_i\in\S_+^{\uparrow}[1, k_1]}\sum_{\x_j\in\S^\downarrow_-[1, k_2]}\ell(f_\bw (\x_j) - f_\bw(\x_i)). 
\end{equation} 
Regarding the pairwise surrogate loss  $\ell(f_\w(\x_-) - f_\w(\x_+))$, different choices have been investigated in the literature. A list of different surrogate loss functions with a sampling of references are summarized in Table~\ref{tab:loss}. An important property of pairwise surrogate loss for AUC maximization is its consistency~\cite{10.5555/2832249.2832379}. Loosely speaking,  a pairwise surrogate loss $\ell(\cdot)$ is called consistent if optimizing the surrogate loss with infinite amount of data  gives a solution to optimizing the original AUC score. A rigorous definition is given in~\cite{10.5555/2832249.2832379}. A necessary condition for the consistency of a pairwise loss $\ell(f(\x') - f(\x))$   for a positive-negative pair $(\x, \x')$ is that $\ell$ is a convex, differentiable and non-decreasing function with $\ell'(0) > 0$~\cite{10.5555/2832249.2832379}. 

\begin{table*}[t] 
	\caption{Different Surrogate Loss Functions $\ell(s)$ used in AUC maximization}\label{tab:0} \vspace*{-0.1in}
	\centering
	\label{tab:loss}
	\scalebox{0.72}{\begin{tabular}{lllcc}
			\toprule
	        Name&Form&Parameters&Remarks&References\\
		    \hline 
		    Square&$\ell(s)=(s+c)^2$&$c>0$&Consistent&\cite{gao2013one,ying2016stochastic,ding2017large,shi2020quadruply,liu2019adaptive}\\
			Hinge&$\ell(s)=\max(0,c+s)$&$c>0$&Non-consistent&\cite{brefeld2005auc,khalid2018scalable,pmlr-v28-kar13,wang2013online}\\
			Squared Hinge&$\ell(s)=max(0, c+s)^2$&$c>0$&Consistent&\cite{10.5555/3041838.3041945,brefeld2005auc,khalid2018scalable}\\
			Logistic&$\ell(s)=-\log \frac{1}{1+\exp(cs)}$&$c>0$&Consistent&\cite{sulam2017maximizing}\\
			Exponential Loss&$\ell(s) = \exp(cs)$&$c>0$&Consistent&\cite{10.5555/945365.964285}\\
			\midrule
			Barrier Hinge&$\ell(s) = \max(-b(r+s)+r, \max(b(s-r), r-s))$ &$r>0, b>0$&Noisy Labels&\cite{charoenphakdee2019symmetric}\\
			\midrule
		    Sigmoid& $\ell(s) = \frac{1}{1+\exp(-cs)}$ &$c>0$&Non-convex&\cite{calders2007efficient,herschtal2004optimising,Iwata2020SemiSupervisedLF,Ueda2018PartialAM,Wang2011MarkerSV}\\
		    Ramp Function&$\ell(s)=\max(0, 1+s) - \max(0, c+s)$&$c<0$&Non-convex&\cite{cheng2018adaptive}\\
		    CDF of Normal Distribution&$\ell(s)=\frac{1}{2}(1+erf(\frac{s}{\sqrt{2}c}))$&$c>0$&Non-convex&\cite{Komori2010ABM}\\
		    Exponential-type&$\ell(s) = 1 - \exp(-cs)$&$c>0$&Non-convex&\cite{takashi12}\\
		    \midrule
		    Chebyshev Polynomial&$\ell(s)=\sum_{k=0}^mc_ks^k$&$m>0$&Decomposable&\cite{calders2007efficient}\\
		    \bottomrule
	\end{tabular}}
	\vspace*{0.1in} 
\end{table*}

\paragraph{\bf Min-Max Objectives for AUC Maximization.} One issue of the pairwise loss based objective is that it needs to explicitly construct the positive-negative pairs, which is not suitable for online learning where the data comes sequentially and distributed optimization where data is distributed over many different machines. To address this issue,  Ying et al.~\cite{ying2016stochastic} propose to formulate an equivalent min-max objective  for using a {\bf pairwise square loss}. Specially, when $\ell(f_\bw (\x_j) - f_\bw(\x_i)) = (c+f_\bw (\x_j) - f_\bw(\x_i))^2$, the problem in~(\ref{eq:auc-emp}) is equivalent to
	\begin{align}\label{opt:spp}
	\min_{\w\in\R^d,(a,b)\in\R^2}\max_{\alpha\in\R}F\left(\w,a,b,\alpha\right):=\EX_{\z}\left[F\left(\w,a,b,\alpha;\z\right)\right],
	\end{align}
where $\EX_{\z}$ denotes the empirical average of training data (offline setting) or expectation over underlying distribution (online setting),   $F(\w, a, b, \alpha; \z)$ is given by 
\begin{align}\label{eqn:AUCF}	F(\w,a,b,\alpha;\z)=&(1-p)\left(f_\w(\x)-a\right)^2\mathbb{I}(y=1)+p(f_\w(\x)-b)^2\mathbb{I}(y=-1)\\&-p(1-p)\alpha^2+2\alpha\left(p(1-p)c+ p f_\w(\x)\mathbb{I}(y=-1)-(1-p)f_\w(\x)\mathbb{I}(y=1)\right)\notag, 
\end{align} and $p = \Pr(y=1)$ (online setting) or $p=n_+/n$ (offline setting).  
A benefit of this objective function is that it is decomposable over individual examples.  Hence it enables  one to develop efficient  stochastic algorithms for updating the model parameter $\w$ without explicitly constructing and handling positive-negative pairs. It is notable that a similar min-max formulation for AUC maximization was also independently examined in~ \cite{DBLP:conf/nips/PalaniappanB16} for the offline setting. 

Recently, Yuan et al.~\cite{DBLP:journals/corr/abs-2012-03173} reveal some potential issues of optimizing pairwise square loss and its equivalent min-max objective. They demonstrate that optimizing the pairwise square loss or its equivalent min-max objective is sensitive to noisy data and also has adverse effect on easy data. To address these issues, they decompose the square loss based objective into three components:  
\begin{align}\label{eqn:sq}
 &\EX[(c - h_\w(\x) +  h_\w(\x'))^2|y=1, y'=-1]\\
 &= {\EX[(h_\w(\x) - a(\w))^2|y=1]}+  {\EX[(h_\w(\x') - b(\w))^2|y'=1]}+  {(c - a(\w) +  b(\w))^2}\notag,
\end{align}
where $a(\w)=\EX_{\x}[f_\w(\x)|y=1]$ and $b(\w) = \EX_{\x}[f_\w(\x)|y=-1]$, and they propose to replace the last term by using a squared hinge loss: 
 \begin{align}\label{eqn:AUCM}
\min_{\w\in\R^d}{\EX[(h_\w(\x) - a(\w))^2|y=1]}+  {\EX[(h_\w(\x') - b(\w))^2|y'=1]}+  (c- a(\w) +  b(\w))_+^2,
\end{align}
whose objective is referred to as min-max margin loss~\cite{DBLP:journals/corr/abs-2012-03173}.  For solving the above problem, they formulate the problem into an equivalent min-max optimization problem: 
	\begin{align}\label{opt:spp2}
	\min_{\w\in\R^d,(a,b)\in\R^2}\max_{\alpha\geq 0}f\left(\w,a,b,\alpha\right):=\EX_{\z}\left[F\left(\w,a,b,\alpha;\z\right)\right],
	\end{align}
	where $F(\w, a, b,\alpha; \z)$ is the same as~(\ref{eqn:AUCF}). The difference between the above objective and~(\ref{opt:spp}) is that there is a non-negative constraint on the dual variable $\alpha\geq 0$. 

\paragraph{\bf Composite Objectives for AUC Maximization.} Recently, Zhu et al.~\cite{zhubenchmark} propose another family of objectives for AUC maximization, which subsumes min-max objective of the pairwise square loss and the min-max margin loss as special cases. The objective consists of three terms: 
	\begin{align}\label{opt:comps}
\min_{\w\in\R^d}{\EX[(f_\w(\x) - a(\w))^2|y=1]}+  {\EX[(f_\w(\x') - b(\w))^2|y'=1]}+  \ell(c- a(\w) +  b(\w))^2,
	\end{align}
	where $\ell(\cdot)$ is a  surrogate loss. When $\ell(\cdot)$ is a square function, the above objective is equivalent to the pairwise square loss based objective or the min-max objective in~(\ref{opt:spp}). When $\ell(\cdot)$ is a squared hinge function, the above objective is equivalent to the min-max margin objective~(\ref{opt:spp2}). Other choices of $\ell(\cdot)$ are possible~\cite{zhubenchmark}. For solving the above objective, it can be transformed into
	\begin{align}\label{opt:obj3}
\min_{\w\in\R^d, (a, b)\in\R^2}{\EX[(f_\w(\x) - a)^2|y=1]}+  {\EX[(f_\w(\x') - b)^2|y'=1]}+  \ell(c- a(\w) +  b(\w))^2,
	\end{align}
	where the last term can be regarded as a two-level stochastic compositional function~\cite{wang2017scgd,ghadimi2020nasa}. Another way is to view all three terms in~(\ref{opt:comps}) as compositional functions. 
	
	It is notable that a regularization term about $\w$ (e.g., $\ell_2$ norm square) can be added to the above objectives for improving generalization. {In addition, formulations can be extended to the multi-class scenario following the one-vs-all or one-vs-one settings~\cite{9502525,10.1023/A:1010920819831,liu2019stochastic}.}

\section{Full Batch Based Methods - The First Age} \label{sec:full}
Earlier works for AUC maximization use full batch based methods, which process all training examples at each iteration in the algorithmic optimization.  Notable  optimization algorithms for AUC maximization include the quadratic programming, gradient decent methods, cutting plane algorithms, and  boosting-type methods. 

\subsection{AUC Maximization}
\paragraph{\bf Quadratic Programming.}  To the best of our knowledge, the earliest work dates back to 1999~\cite{Herbrich1999d} which derives the dual problem of the support vector machine (SVM) formulation for ordinal regression in the kernel setting. Quadratic programming  is then employed to obtain the optimal solution which can apply to AUC maximization.  The work  \cite{brefeld2005auc,Rakotomamonjy04supportvector} use a  similar optimization algorithm for AUC maximization with the hinge loss.  Since the number of constraints and parameters  grows 
quadratically in the number of examples, running such quadratic programming for AUC maximization  is very computationally expensive for large-scale datasets.  To mitigate such computational burden, in \cite{brefeld2005auc,Rakotomamonjy04supportvector} heuristic tricks using k-means clustering and k-nearest neighborhood are proposed to reduce the number of constraints. However, such approximate solutions do not guarantee an optimal solution to the original AUC maximization problem.

\paragraph{\bf Gradient Descent Methods.} Gradient descent methods are used in  \cite{calders2007efficient,10.5555/3041838.3041945,herschtal2004optimising} for AUC maximization. \cite{10.5555/3041838.3041945} is probably the first work that applies the gradient descent method for AUC maximization. They use the the hinge function with a power $p>1$ as the surrogate loss. One year later, the work \cite{herschtal2004optimising} considers improving the gradient descent algorithm for AUC maximization, where they use the sigmoid function as a surrogate loss.  They also propose a heuristic technique by reducing the number of positive-negative pairs  used in the gradient descent methods.  In particular, for each negative data they only construct  a pairwise loss with only one positive data. However, the quality of  such approximation highly depends on the properties of the dataset. When the examples have large intra-variance, their objective could yield poor performance. The work \cite{calders2007efficient}  uses a different method to improve the scalability of the gradient descent method. In particular, they use the Chebyshev polynomial to approximate the indicator function in the original formulation of the AUC score given by \eqref{eqn:eaucd} and then a gradient descent method is employed to optimize such approximated AUC score, which only requires a linear scan of all examples at each iteration without explicitly working on all pairs.  

\paragraph{\bf Cutting Plane and Accelerated Gradient-Based Methods.} The seminal work by  \citet{Joachims}  uses the cutting plane algorithms to optimize  a general multivariate performance measure including the AUC score. The basic principle behind this optimization algorithm is to, at each iteration,  solve a  quadratic programming problem subject to a selected subset of constraints. The sufficient subset of constraints is generated by  gradually adding the
currently most violated constraint in each iteration.   The cutting plane methods converge with an iteration complexity of  $\mathcal{O}(\frac{1}{\lambda\epsilon})$ to find a $\epsilon$-accurate solution, where $\lambda$ is the regularization parameter in the formulation.  \citet{Xinhua} work on the dual form of  the formulation for optimizing the multivariate performance measure, which may be not smooth, and use the smoothing techniques \cite{nesterov2005smooth} to smooth the empirical objective function. Then,  the Nesterov’s accelerated gradient method \cite{nesterov1983method} is employed to optimize the smoothed objective function, which has an iteration complexity of  $\max (\mathcal{O}(\frac{1}{\epsilon},  \frac{1}{\sqrt{\lambda\epsilon}}))$.

\paragraph{\bf Boosting Methods.}
 Freund et al.~\cite{10.5555/945365.964285}  propose a boosting method named RankBoost for  bipartite ranking, which is applicable to AUC maximization.  The RankBoost algorithm is
based on Freund and Schapire’s AdaBoost algorithm~\cite{10.5555/646943.712093} and its  successor developed by Schapire and Singer~\cite{10.1145/279943.279960}. RankBoost works by combining many ``weak" rankings of the given instances to learn a strong ranking model. The RankBoost algorithm was later applied to AUC maximization~\cite{Cortes2003AUCOV}. The boosting methods for AUC maximization have also been examined in~\cite{NIPS2007_c5ff2543}.

\subsection{Partial AUC Maximization}
 Compared to AUC maximization, partial AUC maximization is much more challenging due to that it involves selection of examples whose prediction scores are in a certain range. We provide a survey of partial AUC maximization according to the chronological order and group them according to the underlying methodologies.

\paragraph{\bf Indirect Methods.} Wu et al.~\cite{10.1145/1401890.1401980} propose a new support vector machine (SVM) named assymetric SVM, which aims to lower the false positive rate while maximizing the margin.  To achieve this, it maximizes two margins, the core-margin (i.e., the margin between the negative class and the high confidence subset of the positive class), and the traditional class-margin. By enlarging the core-margin, it is able to enclose the core (i.e., high confident examples) of the positive class in a set. The authors employ Sequential Minimal Optimization (SMO) to solve the resulting objective.  

Rudin~\cite{JMLR:v10:rudin09b} proposes the p-norm push method for bipartitie ranking, which is to optimize a measure focusing on the left end of the ROC curve aiming to the make the leftmost portion of ROC curve higher. The measure to be minimized is defined as a sum of p-norm of the heights of negative examples, where the height of a negative example is defined as the number of positive examples that are ranked lower than the negative example. The author proposes a boosting-type algorithm for optimizing the p-norm push objective. 

Later, Agarwal~\cite{Agarwal2011TheIP} proposes the infinite-push method  to minimize the maximal height of all negative examples, which can be considered as an empirical estimator of pAUC with the FPR controlled below $1/n_-$. The author proposes a gradient descent algorithm for solving the infinite-push objective, which suffers a higher per-iteration cost in the order of $O(n_+n_-d+n_+n_-\log(n_+n_-))$ and an iteration complexity of $O(1/\epsilon^2)$, where $d$ is the dimensionality of input data.  

Rakotomamonjy~\cite{Rakotomamonjy2012SparseSV} extends the infinite-push method to handle sparsity-inducing regularizers and proposes an ADMM-based algorithm for optimizing the problem, which has a per-iteration cost of $O(n_+n_-d + n_+n_-\log (n_+n_-))$ and an iteration complexity of $O(1/\epsilon)$. In 2014, Li, Jin and Zhou~\cite{Li2014TopRO} propose a method called TopPush for optimizing the infinity-push objective. The authors use a different formulation from that in~\cite{Agarwal2011TheIP} where each  positive example is only compared with the negative example with the highest score before computing the loss, which leads to a more efficient algorithm with a per-iteration cost of $O((n_+ + n_-)d)$. They employ the Nesterov's accelerated gradient method to optimize the dual objective with an iteration complexity of $O(1/\sqrt{\epsilon})$.

\paragraph{\bf Boosting-type methods.} Komori and Eguchi~\cite{Komori2010ABM} propose a boosting-style algorithm named pAUCboost for partial AUC maximization. In this work, the indicator function $\I(f(\x_-)>f(\x_+))$ is approximated by the cumulative density function of the normal distribution.  The weak leaner is defined by a natural cubic spline. To simplify the optimization for the weaker learner and its weight at each iteration, the algorithm first employs one-step Newton-Raphson to update the weight and then solves for the optimal weaker learner given its weight. However, it does not discuss complexity and efficiency in finding weaker learners for maximizing pAUC at each iteration. Takenouchi and Komori and Eguchi~\cite{takashi12} propose a more principled boosting method for pAUC maximization named pU-AUCBoost, where U stands for a surrogate function of the indicator $\I(f(\x_+)>f(\x_-))$. To address the inter-dependency issue between the weak learner and its weights, they derive a lower bound of the pAUC objective at each iteration, which decouples the weaker learner and its weights. In these papers, the authors only conduct experiments on small scale datasets with few hundred or thousand examples. 

\paragraph{\bf Heuristic Methods.} Wang and Chang~\cite{Wang2011MarkerSV} consider the marker (feature) selection  problem  via maximizing the partial AUC of linear risk scores. They propose a surrogate loss function for pAUC and show its non-asymptotic convergence and greedily select features for learning a linear classifier. There is no discussion on efficiency and complexity of how to solve the pAUC maximization problem. The authors have conducted experiments on some simulated data and real data with only few hundred examples.  Ricamato and  Tortorella~\cite{Ricamato2011PartialAM} examine the problem  of how to combine two or multiple classifiers to maximize partial AUC. The problem is reduced to optimizing a scalar combination weight, which is different from standard pAUC maximization methods for learning a classifier. For combining multiple classifiers, they use a greedy method to select which two classifiers to combine at each iteration. As a result, they derive a boosting algorithm similar to the classical Adaboost algorithm, which first finds the optimal base learner given previous combined learner and then optimizes the weight of the base learner. 

\paragraph{\bf Structural SVM Methods.} Narasimhan and Agarwal~\cite{pmlr-v28-narasimhan13} propose a structural SVM based approach for learning a linear model by optimizing partial AUC inspired by~\cite{Joachims}. Their formulated optimization problem has an exponential number of constraints, one for each possible ordering of training data. To solve this problem, they use the cutting plane
method, which is based on the fact that for any $\epsilon>0$ a small subset of the constraints is sufficient to find an $\epsilon$-approximate solution to the problem. However, the bottleneck lies at finding the most violated constraint at each iteration, which could cost $O((n_++n_-)d+n_+n_-+ n_-\log n_-)$ time complexity.   In addition,  the cutting-plane method could have a slow convergence with an iteration complexity of $O(1/\epsilon)$. In the extended version~\cite{Narasimhan2017SupportVA}, the authors have managed to reduce the per-iteration time complexity  to $O((n_++n_-)d+n_+n_-\beta+ n_-\log n_-)$, where $\beta\in(0,1)$ is the upper bound parameter of the FPR. In 2013, the same authors propose a tight surrogate loss for the partial AUC in the structural SVM framework~\cite{10.1145/2487575.2487674}. In this paper, the authors also present a projected gradient method, which suffers a per-iteration cost of $O((n_++n_-)d+ n_-\log n_- + (n_++n_-\beta)\log(n_++n_-\beta))$ for learning a linear model of dimentionality of $d$, and an iteration complexity of $O(1/\epsilon^2)$. A DC programming approach is also presented in~\cite{Narasimhan2017SupportVA} for optimizing pAUC with FPR restricted in a range $(\alpha_0, \alpha_1)$ where $\alpha_0>0$, which is computationally more expensive than the  structural SVM approach due to requiring to solve an entire structural SVM optimization at  each iteration. In these papers, the authors have conducted experiments on multiple datasets with size ranging from a few thousand to a few hundred thousand. {The theoretical work 
\cite{maurer2020} provides a statistical performance guarantee for algorithms of maximizing the empirical pAUC proposed in \cite{pmlr-v28-narasimhan13,Narasimhan2017SupportVA,10.1145/2487575.2487674}.}


\begin{table*}[t] 
	\caption{Comparison of different studies for pAUC maximization.  
}\label{tab:0} \vspace*{-0.1in}
	\centering
	\label{tab:pAUC}
	\scalebox{0.72}{\begin{tabular}{lllcccc}
			\toprule
	        Work	&Category&Objective Functions&Models &\makecell{Complexity/Convergence\\ Analysis}&\makecell{Size of Data}\\
		    \hline 
			\cite{10.1145/1401890.1401980}&Indirect Methods&SVM-like&Kernel&No&$10^5$\\
			\cite{JMLR:v10:rudin09b}&Indirect Methods&P-norm Push&Linear&No&$10^4$\\
			\cite{Agarwal2011TheIP}&Indirect Methods&Infinity Push&Linear&Yes&$10^3$\\
			\cite{Rakotomamonjy2012SparseSV}&Indirect Methods&Infinity Push&Linear&Yes&$10^3$\\
			\cite{Li2014TopRO}&Indirect Methods&Infinity Push&Linear&Yes&$10^6$\\
			\midrule
			\cite{Komori2010ABM}&Boosting-type&pAUC Surrogate&Cubic Spline&No&$10^3$\\
			\cite{takashi12}&Boosting-type&pAUC Surrogate&Decision Stump&No&$10^3$\\
		    \midrule
			\cite{Wang2011MarkerSV}&Heuristic Methods&pAUC Surrogate&Linear&No&$10^2$\\
			\cite{Ricamato2011PartialAM}&Heuristic Methods&pAUC Surrogate&Any&No&$10^5$\\
		    \midrule
		    \citep{pmlr-v28-narasimhan13,Narasimhan2017SupportVA,10.1145/2487575.2487674}&Structural SVM&pAUC surrogate&Linear&Yes&$10^6$\\
		    \midrule
		    	  \citep{eban2017scalable,cotter2019two,narasimhan2020approximate,kumar2021implicit}&Constrained Opt.&Riemann approximation&Linear/Non-linear&Convex Only&$10^5$\\
		    \midrule
		    \cite{pmlr-v139-yang21k}&Stochastic/Deep&Appr. Pairwise Surrogate&Deep Nets&No&$10^5$&\\ 
		    \cite{otpaUC}&Stochastic/Deep&(\ref{eq:pauc-emp}), (\ref{eq:tpauc-emp}),~(\ref{eqn:pauceskl}),~(\ref{eqn:tpaucdro})&Deep Nets&Yes&$10^5$&\\ 
		     \cite{paUCyao}&Stochastic/Deep&(\ref{eq:pauc-emp})&Deep Nets&Yes&$10^5$&\\ 
			\bottomrule
	\end{tabular}}
\end{table*}

 {\paragraph{\bf Constrained Optimization} Maximizing the partial AUC can be reformulated as a constrained optimization problem which involves optimizing a non-decomposable evaluation metric with a certain thresholded form, while constraining another metric of interest. In particular,  the work \cite{eban2017scalable} proposes to approximate the area under the ROC curve using a Riemann approximation while dividing the range of FPRs into a number of bins where each threshold is associated with a bin. This approach allows the reformulation of a constrained optimization problem where the objective is to maximize the sum of the TPRs at each threshold with constraints associated with threshold satisfying the FPRs. Replacing  TPRs and FPRs with surrogate relaxations, it can   be further shown to be equivalent to  a Lagrangian (mini-max) problem   and then vanilla  stochastic gradient descent and ascent algorithms can be applied.  The follow-up
work \cite{cotter2019two,narasimhan2020approximate}  have improved this approach    using the surrogate relaxations  for the primal updates. In \cite{kumar2021implicit}, the authors further improved this approach by expressing  the threshold parameter as a function of the model parameters via the Implicit Function theorem \cite{tu2011introduction}. The resulting optimization problem can be solved using standard gradient based methods. }

\subsection{\bf Summary.} 
  We compare different methods in Table~\ref{tab:pAUC} for pAUC maximization from different perspectives, where we also include deep partial AUC maximization methods reviewed in Section~\ref{sec:dam}. 
  The full-batch based algorithms could suffer a quadratic time complexity in the worst-case or a super-linear (e.g. log-linear) time complexity per-iteration, which makes them not amenable for handling  large-scale datasets.  Most of them are for learning traditional models (e.g., linear models, kernel models) and algorithms for solving the underlying optimization problem are not scalable to large-scale datasets and not suitable for  deep learning. 

\section{Online AUC Maximization - The Second Age}\label{sec:onl}
In contrast to the full-batch methods which need all training  data beforehand, online learning algorithms \cite{cesa2006prediction} can update the model parameter upon receiving new datum and can efficiently handle streaming data where examples are presented in sequence. Online learning with point-wise loss has been studied extensively~\cite{hazan2019introduction,orabona2019modern,shalev2011online}. However, online learning for AUC maximization has different challenges due to that the pairwise loss does not naturally fit the streaming data. In the literature, there have been a wave of studies focusing on online learning for AUC maximization. Below, we will categorize them into two classes, namely, {\bf online buffer-based methods,  online statistics-based methods}. Revolving around these methods, we will discuss two theoretical properties, i.e., regret bounds and statistical error bounds.   

We first provide some background on regret bounds and statistical  error bounds.  In the standard online learning setting, there is no statistical assumption on the data received, e.g., IID assumption. Hence, the measure of interest is the regret bound. Let $\{(\x_1, y_1), \ldots, (\x_T, y_T)\}$ denote the sequence of data received in the stream. To measure the regret, let $L_t(\w_t, \x_t, y_t)$ denote the cost measure of the $t$-th model $\w_t$ with respect to the received data $(\x_t, y_t)$ at the $t$-th iteration, let $L(\w, \{\x_t, y_t\}_{t=1}^T)$ denote the cost measure defined on all data.  Then the regret is defined as 
\begin{align*}
R_T =     \sum_{t=1}^TL_t(\w_t, \x_t, y_t) - \min_{\w}\sum_{t=1}^TL_t(\w, \x_t, y_t).
\end{align*}
There are different ways to define the cost at each iteration for AUC maximization, which will be discussed in the following.  

When the received data is assumed to follow the IID assumption, the statistical error bound is another performance guarantee of interest. There are two types of statistical error bounds, namely generalization error bounds and excess risk bounds, where the former refers to the bounds of the difference between the expected risk of a learned model and the empirical risk, and the latter refers to the bounds of the difference between the expected loss of a learned model and the optimal expected loss.   
In particular, let $\mathcal R(\w)$ denote the expected risk for AUC maximization which is given by $\mathcal R(\w)=\EX_{\x_+\sim\P_+, \x_-\sim\P_-}[\ell(f_\w(\x_j) - f_\w(\x_i))]$.   Then the generalization error bounds usually take the form of $\mathcal R(\widehat\w_T)\leq \frac{1}{T}\sum_{t=1}^TL_t(\w_t, \x_t, y_t) + O(T^\alpha)$ for some $\widehat\w_T$ and $\alpha>0$, and the excess risk bounds usually take the form of $\mathcal R(\widehat\w_T)\leq \min_{\w\in\mathcal W}\mathcal R(\w) + O(T^{-\alpha})$ for some $\widehat\w_T$ and $\alpha$.

\subsection{Online Buffered Gradient Descent for AUC Maximization.} The most representative online buffer-based methods is the online buffer gradient descent method proposed in  the seminal paper \cite{Zhaoicml11} in 2011 by  Zhao, Hoi, Jin and Yang. It is the first work that studies online AUC maximization and inspires many following studies. They propose online buffered gradient descent methods, whose algorithmic framework is shown in Algorithm~\ref{alg:online}. There are two key functions, i.e., $\text{UpdateBuffer}$ and $ \text{UpdateModel}$. In the paper, the authors define the following cost function for each iteration: 
\begin{align*}
    L_t(\w, \x_t, y_t) =  \sum_{i<t, y_i=1}\I(y_t=-1)\ell(f_\w(\x_t) - f_\w(\x_i)) +  \sum_{i<t, y_i=-1}\I(y_t=1)\ell(f_\w(\x_i) - f_\w(\x_t)).
 \end{align*}
They update the buffer by using the ``reservoir sampling” technique~\cite{vitter1985random}, which aims to simulate a uniform sampling of the received examples. They update the model parameter based on the gradient descent of the cost function $L_t$ by only using examples in the buffer, i.e., $(\x_i, y_i)\in\mathcal B_t$. They establish a regret bound in the order of $\sqrt{B_+T_+^3 + B_-T_-^3}$, where $B_+$ and $B_-$ denote the buffer size for positive samples and negative samples, respectively, $T_+$ and $T_-$ denote the number of received positive examples and negative examples over $T$ iterations, respectively. The authors provide an explanation regarding the optimal buffer size in the presence of the variance terms that have been ignored in the regret bound analysis, which gives an optimal buffer size $B_+=\sqrt{T_+}$ and $B_-=\sqrt{T_-}$. 

Later, the statistical error  bounds of  online buffer-based methods are established in~\cite{wang2012generalization,pmlr-v28-kar13}. Wang et al.~\cite{wang2012generalization} provide the generalization error bounds for any arbitrary online learner with an infinite buffer size and a finite buffer size for learning from $n$ examples.  They use the covering number to bound the complexity of hypothesis and derive a generalization error bound of a tailed-averaged solution in the order of $O(\log(\mathcal N(\epsilon) n/\delta)/\sqrt{\min(T, B)})$ with a high probability $1-\delta$, where $\mathcal N(\epsilon)$ denotes the cardinality of $\epsilon$-net of the hypothesis space for a small value $\epsilon$, and $B$ denotes the buffer size. Kar et al.~\cite{pmlr-v28-kar13} improve the generalization error bound of the method with an infinite buffer by using the Rademacher complexity of the hypothesis space. Their error bound of the averaged solution is in the order of $O(C_d/\sqrt{T})$, where $C_d$ is a constant in the Rademacher complexity that is dependent only on the dimension $d$ of the input space. Based on the generalization error bound and the regret bound, they also establish  an excess risk bound in the order of $R_T/T  + O(\frac{C_d+\log(T/\delta)}{T})$, where $R_T$ is the regret bound and $\delta\in(0,1)$ is the failure probability. In addition, they also provide generalization error bounds and a high probability excess risk bound for online buffered gradient descent method with a finite-sized buffer, which has a dominating term of $O(C_d\log(T/\delta)/\sqrt{B})$, where $B$ is the buffer size. 

\begin{algorithm}[t]
\caption {Online Buffered Gradient Descent Method  for AUC Maximization} \label{alg:online}
\begin{algorithmic}[1]
\STATE{Initialization: $\w_0\in \R^d$, $\mathcal B_0=\emptyset$,}
\FOR{$t=1,2, ..., T$}
\STATE Receive a data $\{\x_t, y_t\}$
\STATE Update the Buffer $\mathcal B_t=\text{UpdateBuffer}(\mathcal B_{t-1}, \x_t, y_t)$
\STATE Update the model parameter $\w_{t} = \text{UpdateModel}(L_t, \w_{t-1}, \mathcal B_t, \x_t, y_t)$
\ENDFOR 
\end{algorithmic}
\label{alg:main}
\end{algorithm}

Kar et al.~\cite{10.5555/2968826.2968904} also study an online buffer-based method with an infinite buffer size for partial AUC maximization. In the paper, they define a different cost function for each iteration. Let $L(\w, \{\x_i, y_i\}_{i=1}^t)$ denote the pairwise loss summed over all pairs received in the first $t$ iterations. The cost function at the $t$-th iteration is defined as $L_t(\w; \x_t, y_t) = L(\w, \{\x_i, y_i\}_{i=1}^t) - L(\w, \{\x_i, y_i\}_{i=1}^{t-1})$. In this way, the optimal model in hindsight $\min_{\w}L_t(\w; \x_t, y_t)$ indeed optimizes the objective of interest (e.g., pairwise-loss based objective for AUC maximization). They employ the Follow-the-Regularized-Leader (FTRL) algorithm for updating the model and establish a regret bound in the order of $\sqrt{T_+T_-^2 + T_-T_+^2}$. They also establish an excess risk bound for a modified FTRL method which uses $s$ samples per-iteration, which is in the order of $O(1/T^{1/4})$ for $s=\sqrt{T}$.


\subsection{Online Statistics-Based Methods.} To address the issue of maintaining a large buffer size, Gao et al.~\cite{gao2013one} propose an online AUC maximization by leveraging the property of pairwise square loss for learning a linear model. They use the same definition of the cost function $L_t(\w; \x_t, y_t)$ as~\cite{Zhaoicml11}. By using the square loss for learning a linear model, they show that the gradient of the cost function $L_t$ can be computed based on first-order moments (mean vectors of positive and negative examples) and second-order moments (covariance matrices of positive and negative examples) of the received data before the $t$-th iteration. Nevertheless, it also introduces high memory costs for maintaining the covariance matrix. To address this issue, the authors develop low-rank approximation methods and only update  low-rank matrices for the second-order moments at each iteration. In the paper, the authors also establish the regret bounds for both full-rank and the low-rank approximation methods.

\subsection{Online Non-linear Methods for AUC maximization.} Online nonlinear kernel methods based on AUC maximization have been proposed and studied in \cite{ding2017large,hu2017online,szorenyi2017non} to address the non-separability of the data and the  scalability issues. In particular, \citet{ding2017large} extend the online buffered gradient descent method to learn non-linear kernel-based models. They employ two functional approximation strategies, i.e.,  random fourier features (RFF) \cite{rahimi2007random} to approximate the shift invariant kernels and the Nystr\"{o}m method \cite{williams2001using}  to approximate the kernel matrix. For the two methods, the authors have established regret bounds in the order of $\sqrt{T}$. Nevertheless, it is claimed that the RFF based method require $m=T$ random features for achieving a high probability bound.

Hu et al. (2017)~\cite{hu2017online} propose a different kernelized  online AUC maximization method. They do not use RFF or the Nystr\"{o}m method to approximately compute the kernel similarities. Instead, they use the pairwise hinge loss or squared hinge loss as the surrogate loss, and maintain support vectors of positive and negative classes in the online fashion, i.e., those examples whose contribution weights in the classifier are non-zero. They maintain and update two buffers for storing these support vectors and their contribution weights. The cost function at each iteration is defined similarly as in~\cite{Zhaoicml11} except that $k$-nearest examples to the received data in the buffer are used to compute the loss.  They establish a regret bound in the order of $\sqrt{T}$.  They also present an extension to the multiple kernel learning framework which can automatically determine a good kernel representation. 

\citet{szorenyi2017non} propose a kNN-based online AUC maximization method by suggesting an algorithmic solution based on the kNN-estimate of the conditional probability function. They use an infinite buffer that stores all received examples.

\subsection{Adaptive Online AUC Maximization.}  The work  \cite{liu2019adaptive,cheng2018adaptive,ding2015adaptive} propose and study adaptive online AUC maximization algorithms belonging to the two classes of online AUC maximization methods.  Ding et al.\cite{ding2015adaptive}  extend the online statistics-based method proposed in~\cite{gao2013one} for AUC maximization by incorporating an online adaptive gradient method (AdaGrad) \cite{duchi2011adaptive} for exploiting the knowledge of historical gradients.  Cheng et al. \cite{cheng2018adaptive} propose to use the Adam-style update in the framework of online buffered gradient descent methods, where the buffer is maintained in first-in-first-out fashion. In the paper, they use a non-convex ramp loss as the surrogate function of the indicator $\I(f_\w(\x_-) - f_\w(\x_+))$ and use concave-convex procedure (CCCP) to approximate the cost function at each iteration.  Liu et al. \cite{liu2019adaptive} leverage the Adam-style update~\cite{kingma2014adam} in the framework of online statistics-based method for AUC maximization. 

\vspace*{-0.1in}
\subsection{Summary} Two classes of methods namely online buffer-based methods and online statistics-based methods have been proposed for online AUC maximization. Online buffer-based methods are more generic, which can be leveraged for learning both linear and non-linear classifiers for any possible pairwise surrogate losses, while online statistics-based methods are restricted to learning linear models and using pairwise square loss. Nevertheless, online buffer-based methods usually require a large buffer to achieve a good performance, and online statistics-based methods could enjoy a lower regret and a lower memory costs for low-dimensional data.  We compare different works in Table~\ref{table:online} from different perspectives.

\begin{table*}[t] 
	\caption{Comparison of different studies for online AUC maximization, where $T$ is the total number of iterations, $B$ is a fixed buffer size, $r$ is a constant size of the low rank approximation, $d$ is the dimensionality of input data.    	
}\label{table:online} \vspace*{-0.1in}
	\centering
	\scalebox{0.8}{\begin{tabular}{lllcccc}
			\toprule
	        Work	&Category&Loss Functions&Models &\makecell{Regret/Generalization\\ Analysis}&\makecell{Memory Costs}\\
		    \hline 
			\cite{Zhaoicml11}&Buffer-based&Pairwise AUC Surrogate&Linear&Yes&$O(Bd)$\\
			\cite{wang2012generalization}&Buffer-based&Pairwise  AUC Surrogate&Linear&Yes&$O(Td)$\\
			\cite{pmlr-v28-kar13}&Buffer-based&Pairwise AUC Surrogate&Linear&Yes&$O(Td)$ or $O(Bd)$\\
			\cite{10.5555/2968826.2968904}&Buffer-based&Pairwise  pAUC Surrogate&Linear&Yes&$O(Td)$\\
			\cite{ding2017large}&Buffer-based&Pairwise  AUC Surrogate&Kernel&Yes&$O(Bd)$\\
			\cite{hu2017online}&Buffer-based&Pairwise (squared) hinge loss&Kernel&Yes&$O(Bd)$\\
			\cite{cheng2018adaptive}&Buffer-based&Pairwise Ramp loss&Linear&Yes&$O(Bd)$\\
			\cite{szorenyi2017non}&Buffer-based&NA&Non-Parametric&Yes&$O(Td)$\\
			\midrule
			\cite{gao2013one}&Statistics-based&Pairwise square loss&Linear&Yes&$O(d^2)$ or $O(dr)$\\
			\cite{ding2015adaptive}&Statistics-based&Pairwise  square loss&Linear&Yes&$O(d^2)$\\
			\cite{liu2019adaptive}&Statistics-based&Pairwise square loss&Linear&Yes&$O(d^2)$\\
	\bottomrule
	\end{tabular}}

\end{table*}

\section{Stochastic AUC Maximization - The Third Age} \label{sec:sto}
Stochastic AUC maximization refers to a family of methods that only process one or a small mini-batch of examples at each iteration for updating the model parameters, which are amenable for handling big data. The difference from online AUC maximization is that the IID assumption of data is typically assumed in stochastic AUC maximization.   In this section, we  provide a review on works for learning linear and kernel-based models for AUC maximization, and present a review for deep AUC maximization in Section~\ref{sec:dam}.  We categorize the existing stochastic methods for AUC maximization into two classes, i.e., stochastic batch-based pairwise methods, stochastic primal-dual methods.  The existing works consider two learning settings: online setting similar to stochastic approximation in conventional literature~\cite{10.5555/2678054}, and offline setting similar to stochastic average approximation in conventional literature~\cite{10.1137/070704277}.  In the online setting, the data $\{\z_1,\z_2,\ldots, \z_t, \ldots\}$ is assumed to be i.i.d. from an unknown distribution and continuously arriving, i.e., streaming data, and the goal is to minimize the expected loss in~(\ref{eqn:aucd2}).  In the offline setting, a set of training data $\S=\{(\x_i, y_i), i\in[n]\}$ of size $n$ is given beforehand, and the goal is to minimize the empirical loss in~(\ref{eq:auc-emp}). There are two different errors that have been analyzed for different algorithms, namely optimization error and statistical error. For optimizing the expected loss~(\ref{eqn:aucd2}), the optimization error and the statistical error coincides.  

\subsection{Stochastic Batch-based Pairwise  Methods.} The idea of batch-based pairwise methods is to use a mini-batch of data points for computing a stochastic gradient estimator for updating the model parameter. Below, we discuss two categories of methods for the offline setting and the online setting, respectively.


\paragraph{\bf Offline setting.}  A straightforward approach for designing stochastic AUC maximization algorithms is by using  stochastic gradients of  the pairwise loss function $\ell(\bw; \z, \z') = \ell(f_\w(\z') - f_\w(\z))$ for the sampled  positive-negative pairs $(\z, \z')$.  Then the model parameter can be updated by any suitable stochastic algorithms, e.g., SGD. This approach has been adopted and studied in several papers with different aims~\cite{gu2019scalable,dang2020large,lei2021generalization,yang2021simple}. 

\citet{gu2019scalable} focus on establishing statistical error in the order of $O(1/\sqrt{n})$ of a stochastic algorithm based on a finite training data set of size $n$. They propose a doubly stochastic gradient algorithm (AdaDSG) by solving regularized  pairwise learning problems. Specifically, at each stage, AdaDSG uses an inner solver to solve a sampled sub-problem, and then uses the solution obtained from this sub-problem as a warm start for the next larger problem with a doubled size of training samples.  The inner solver simply uses the SGD method based on a randomly sampled positive-negative pair for updating the model parameter. The work \cite{dang2020large} proposes a triply stochastic functional gradient for  AUC maximization problem for learning a kernelized model. At each iteration, this algorithm  performs SGD update based on an unbiased functional gradient calculated from a random pair of examples using random Fourier features (the pair of examples and the random variable for constructing the Fourier features constitute the triplet). A convergence rate in the order $\mathcal{O}({1\over{T}})$ for the optimization error was established for the strongly regularized empirical AUC maximization problem. The work  \cite{shi2020quadruply} also considers a similar algorithm for semi-supervised ordinal regression based on  AUC optimization. 

Recent work \cite{lei2020sharper,lei2021generalization} focus on establishing the statistical error for a specific type of SGD algorithms for pairwise learning in the offline setting. In particular, they study the following SGD-type algorithm for pairwise learning:  at the each iteration, it randomly draws $(\z_{i_t}, \z_{j_{t}})$ from all possible ${n(n-1)/2}$ pairs of examples, and the model parameter is updated  by  $\bw_{t+1} = \bw_{t} - \eta_t \nabla\ell(\bw_{t}; \z_{i_t}, \z_{j_{t}})$.  The work \cite{lei2020sharper} uses the the concept of uniform stability \cite{bousquet2002stability} and the corresponding high probability  generalization bounds  \cite{bousquet2020sharper,feldman2018generalization} to derive the excess risk bound  $\mathcal{O}({\log n/\sqrt{n}})$ in the convex case.   The work \cite{lei2021generalization} further provides improved   results    by incorporating the variance information and show that, under an interpolation or a low noise assumption, the risk bounds can achieve   $\O({1/n})$ through exploiting the smoothness assumption. 

\citet{yang2021simple} proposes a simple SGD-type algorithm for pairwise learning where, at the each iteration, it randomly draws $i_t\in [n]$ and the current example $\z_{i_t}$ is paired with previous one $\z_{i_{t-1}}$, and the model parameter is updated by the SGD based on the pair $(\z_{i_t}, \z_{i_{t-1}})$, i.e., $\bw_{t+1} = \bw_{t} - \eta_t \nabla\ell(\bw_{t}; \z_{i_t}, \z_{i_{t-1}})$.   The authors have established excess risk bounds $\mathcal{O}({1\over \sqrt{n}})$ for smooth and non-smooth convex losses, and smooth non-convex losses under Polyak-Lojasiewicz (PL) condition, in different orders with different number of iterations. 
\paragraph{\bf Online setting.} The {online setting} is more challenging due to that each iteration only receives or samples one data point. The challenge is that an unbiased stochastic gradient cannot be computed based on one data point. To address the challenge, the received data will be stored in a buffer and will be used for computing a stochastic gradient, which is similar to online AUC optimization \cite{wang2012generalization,wang2013online,pmlr-v28-kar13}.  \cite{guo2017online,ying2016online,boissier2016fast} have considered  SGD for pairwise learning in the stochastic setting with an infinite buffer. In particular, for such SGD-based pairwise learning algorithms, at each iteration, the current example $\z_{t}$ is paired with previous ones $\{\z_{1},\ldots,\z_{{t-1}}\}$, the model parameter is updated by gradient descent based on the gradient  of  $L_t(\bw_t; \z_t) = {1\over t-1} \sum_{j=1}^{t-1}  \ell(\bw_t; \z_t, \z_{j})$, i.e., $\bw_{t+1} = \bw_t - \eta_t \nabla L_t(\bw_t; \z_t)$ where $\eta_t$ is a step size.   The authors of \cite{guo2017online,ying2016online} prove the convergence of such SGD-based algorithm  for learning a kernelized model with a  convergence rate  of $\mathcal{O}({1\over t})$ for a strongly convex objective function and   $\mathcal{O}({1\over \sqrt{t}})$ for using the pairwise  square  loss without explicit regularization term.  For learning a linear model, \citet{boissier2016fast} prove that a fast convergence rate $\mathcal{O}({1\over t})$ is still possible for using the pairwise square loss. However, it is notable that the algorithms in \cite{guo2017online,ying2016online,boissier2016fast} are not scalable to large-scale datasets since the buffer size increases as the number of sampled data. 
 This issue can be addressed by the  stochastic primal-dual methods discussed shortly.


\subsection{Stochastic Primal-Dual (PD) Methods.} The idea of stochastic primal-dual methods is to directly apply stochastic methods for addressing the  min-max formulations of AUC maximization, e.g.,~(\ref{opt:spp}).  The benefit of the min-max formulations is that the minimax objective is simply the average of individual data, which makes it suitable to the online setting. Nevertheless, the algorithms discussed below can be applied to both online and offline settings. It is notable that most of the algorithms discussed in this subsection are developed for solving the min-max formulation for the pairwise square loss. However, many of them can be easily extended for solving the min-max margin loss~(\ref{opt:spp2}). 


Ying et al.~\cite{ying2016stochastic} are the first to propose the idea of solving a minimax objective ( i.e., (\ref{opt:spp})) by a stochastic algorithm for AUC maximization with a pairwise square loss. 
The authors propose to use the stochastic first-order primal-dual algorithm \cite{nemirovskistochastic} for AUC maximization, which is referred to as {SOLAM}. The algorithm uses a stochastic gradient descent for updating the primal variables $\w, a, b$ and uses a stochastic gradient ascent for updating the dual variable $\alpha$.  It enjoys a convergence rate $\O({1\over \sqrt{T}})$ with a per-iteration complexity $\O(d)$ for learning a linear model of dimensionality of $d$.  

In the subsequent work \cite{natole2018stochastic,lei2021stochastic}, the authors  leverage the special formulation of the minimax objective for AUC maximization with a square loss to derive faster algorithms for learning a linear model, i.e., $f_\w(\x)=\w^{\top}\x$. In particular, \citet{natole2018stochastic} derive a closed form solution for $a, b, \alpha$ given $\w$ for (\ref{opt:spp}), i.e., $a(\w) = \w^{\top}\EX[\x|y=1]$, $b(\w)=\w^{\top}\EX[\x|y=-1]$ and $\alpha(\w) = b(\w)+ c - a(\w)$~\footnote{In their papers, $\alpha(\w)$ is given by $b(\w)-a(\w)$ due to a variable change.}.  Given that data statistics  $\EX[\x| y=1]$, $\EX[\x| y=-1]$ and the probability $p=\Pr(y=1)$ can easily be estimated from training data, the authors propose a stochastic proximal gradient descent algorithm which only updates $\w$ while the auxiliary variables $a, b$ and $\alpha$  are subsequently computed from $\w$ using the updated data statistics. A fast convergence rate $\O({1\over T})$ is proved for AUC maximization by leveraging the strong  convexity of the regularization term, e.g., $\ell_2$ norm square regularization.  \citet{lei2021stochastic} give an alternative but self-contained proof for stochastic saddle point formulation in \cite{ying2016stochastic,natole2018stochastic} by writing the objective function in \eqref{opt:spp} with $c=1$ as
\begin{align*}\label{eq:decomp0}&(1 - \bw^\top (x - x') )^2   =
\bigl( [1 +  \alpha(\bw) ] + [\bw^\top x' - b(\bw)] - [\bw^\top x - a(\bw)]\bigr)^2 \\
& \!\!= \bigl( \bigl[1 +  \bw^\top (\EX[\tilde{x}|\tilde{y}=-1] - \EX[\tilde{x}|\tilde{y}=1])\bigr] 
 \!+\! \bigl[\bw^\top (x' - \EX [\tilde{x}|\tilde{y}=-1])  - \bw^\top (x - \EX[\tilde{x}|\tilde{y}=1])\bigr]\bigr)^2\numberthis. \end{align*}
Based on this important observation, they prove that AUC maximization \eqref{eq:auc-emp} is equivalent to $ \min_{\bw} \EX_{\z} [  \widetilde{F}(\bw;\z) ] =f(\bw) $,  where \begin{align*}\label{tf}
  & \widetilde{F}(\bw;\z)  =p(1-p)+ (1-p)\big(\bw^\top\big(x-\EX[\x'|y'=1]\big)\big)^2\I(y=1)
  +p\big(\bw^\top\big(\x-\EX[\x'|y'=-1]\big)\big)^2\bI_{[y=-1]} \\
  & +2p(1-p)\bw^\top\big(\EX[\x'|y'=-1]-\EX[\x'|y'=1]\big)
 +p(1-p)\big(\bw^\top\big(\EX[\x'|y'=-1]-\EX[\x'|y'=1]\big)\big)^2.
\end{align*}
From this key observation, they propose a stochastic proximal stochastic gradient (SPAM) which also only needs to update $\bw$. In particular, the authors prove that, in either the unconstrained case without explicit regularizer or with a strong convex regularizer, SPAM can achieve a fast convergence rate $\O({1\over T})$ with a linear per-iteration cost $\O(d).$


There are further studies trying to improve the convergence rate for solving the minimax objective \eqref{opt:spp} without assuming the strong convexity of the regularizer.  Liu et al.~\cite{liu2018fast} propose an improved stochastic algorithm for solving the minimax objective of AUC maximization. The idea is to leverage the strong concavity in terms of the dual variable $\alpha$ and a proved error bound condition of the primal objective function in terms of $\w, a, b$. Their algorithm needs to know the total number of iterations $T$ beforehand and divides the update into multiple stages according to $T$, and each stage calls a stochastic primal-dual method with a constant step size. After each stage, the step size is decreased by a constant factor.  Their algorithm enjoys a convergence rate of $O(1/T)$ for $T$ iterations with one example per-iteration. Later on, Yan et al.~\cite{DBLP:journals/corr/abs-1904-10112} consider a more general minimax objective under an error bound condition of the primal objective and develop a stagewise stochastic algorithm without knowing the total number of iterations $T$ in advance. Their algorithm also enjoy a convergence rate of $O(1/T)$. 

Stochastic algorithms with linear convergence for AUC maximization by using more advanced techniques, e.g., variance-reduction, have been considered in several later works, e.g., \cite{natole2019stochastic,dan2021variance,yang2020-SHT}. \citet{natole2019stochastic} propose a minibatch stochastic primal-dual algorithm (SPDAM)  with a linear convergence rate.  This algorithm is  adapted from the mini-batch stochastic primal-dual coordinate method in \cite{zhang2015stochastic} to the problem of AUC maximization with the pairwise  square loss and a strongly convex regularizer.  The authors prove its  linear convergence rate $e^{ - c(n,m, \lambda) T}$ where $c(n,m,\lambda)$ depends on the size $m$ of the minibatch set, the size $n$ of training data, and the strong convexity parameter $\lambda.$  The work \cite{dan2021variance} further extends SPAM \cite{natole2018stochastic} by using the variance-reduction technique \cite{johnson2013accelerating}. It enjoys a linear convergence rate $e^{-c(M,\beta, \eta)T}$ where $\beta$ is the strongly-convex parameter, $M$ is the strongly smooth parameter and $\eta$ is the constant step size.

In \cite{yang2020-SHT,zhou2020online}, the authors develop efficient sparse AUC maximization algorithms with the pairwise square loss for analyzing the high dimensional data. Both studies use the minimax objective (e.g., \eqref{opt:spp}) and the explicit solutions for the auxiliary variables $a,b$ and $\alpha$ as observed in \cite{natole2018stochastic,lei2021stochastic}. In particular, the work \cite{yang2020-SHT} use the hard thresholding algorithms for AUC maximizatin and prove  its linear convergence    under the assumption of restricted strong convexity (RSC) and restricted strong smoothness (RSS)  on the objective function.  The work \cite{zhou2020online} considers the application of AUC maximization for handling sparse high-dimensional datasets in the sense that the number of nonzero features $k$ in each example is far less than the total number of features $d$.  Such datasets are abundant in online spam filtering \cite{schutte2021using}, ad click prediction \cite{mcmahan2013ad}, and identifying malicious URLs \cite{ma2009identifying}. They develop a generalized Follow-The-Regularized-Leader framework \cite{mcmahan2017survey} for AUC maximization with a lazy update which only involves a per-iteration cost $O(k).$

\begin{table*}[t] 
	\caption{Comparison of different studies for stochastic AUC maximization algorithms, where $T$ is the total number of iterations, $d$ denotes the dimensionality of the input data,  
	$B$ is the batch size used in~\cite{yang2020-SHT} on which the parameter $\rho_k^+(B)$ is dependent,   $m$ is the degree of Bernstein polynomials to approximate the general convex loss  used in \cite{yang2020stochastic},  $\rho_k^+(B)$ and $\rho_k^-$ are RSC and RSS parameters used in \cite{yang2020-SHT}, and $\beta, M$ and $\eta$ respectively denote the strong-convexity parameter, the smooth parameter and the constant step size in \cite{dan2021variance}.  Opt. is short for optimization and Stat. is short for statistical.
}\label{table:stochastic} \vspace*{-0.1in}
	\centering
		\scalebox{0.85}{\begin{tabular}{llccccc}
			\toprule
	        Work	 & Category & Objective & Model &Guarantee  &Rate  &  {Memory Cost}\\
		    \hline 
	\cite{guo2017online} & Stochastic BP	 & Pairwise hinge loss &  kernel&Opt. Error& $\O({1/ T})$ & {$ \O(T^2  )$ }\\
	\cite{ying2016online} & Stochastic BP	 & Pairwise square loss &  kernel &Opt. Error& $\O({1/ {T}^{1/3}})$ &  {$\O(T^2  )$} \\ 
	\cite{boissier2016fast} & Stochastic BP	 & Pairwise square loss &  linear &Opt. Error& $\O({1/ T})$ &  $ \O(d^2 )$ \\ 
	\cite{dang2020large}  & Stochastic BP &Pairwise   loss & linear& Opt. Error& $\O({1/T})$ & $\O(d)$ \\
 	\cite{gu2019scalable}  & Stochastic BP & Pairwise  loss & linear& Stat. Error& $\O({1/{\sqrt{n}}})$ &  $\O(D d)$ \\
	\cite{lei2021generalization} & Stochastic BP	 & Pairwise  loss &  linear & Stat. Error& $\O({1/ \sqrt{n} })$ &  $ \O(d )$ \\
	\cite{yang2021simple} & Stochastic BP	 &Pairwise  loss &  linear &Stat. Error& $\O(1/ \sqrt{n})$ &  $ \O(d)$ \\
	\hline
	\cite{ying2016stochastic} & Stochastic PD	 & Minimax &  linear &Opt. Error& $\O({1/ \sqrt{T}})$ &  $\O(d)$ \\ 
		\cite{liu2018fast} & Stochastic PD  &Minimax (square loss) & linear& Opt. Error& $\O({1/{T}})$ &  $\O(d)$\\
		\cite{DBLP:journals/corr/abs-1904-10112} & Stochastic PD  &Minimax & linear& Opt. Error& $\O({1/{T}})$ &  $\O(d)$\\
		\cite{natole2018stochastic} & Stochastic PD  &Minimax (square loss) & linear& Opt. Error& $\O({1/{T}})$ &  $\O(d)$\\
		\cite{lei2021stochastic}  & Stochastic PD  & Minimax (square loss) & linear&Opt. Error& $\O({1/ {T}})$ &  $\O(d)$\\
		\cite{natole2019stochastic} & Stochastic PD  & Minimax (square loss) & linear&Opt. Error& $\O({1/ {T}})$ &  $\O(d)$\\
		\cite{yang2020-SHT} & Stochastic PD &Minimax (square loss) & linear &Opt. Error& $e^{-c(\rho_k^+(B),\rho_k^- ) T } $  &  $\O(B d)$\\
		\cite{dan2021variance} & Stochastic PD & Minimax (square loss) & linear & Opt. Error& $e^{-c(\beta, \eta, M) T } $  &  $\O(d)$\\
		\cite{yang2020stochastic} & Stochastic PD  & Minimax (general loss) & linear &Opt. Error& $\O({1/\sqrt{m} })$ &  $\O(m d)$\\
	\bottomrule
	\end{tabular}}
\end{table*}

Recently, the work \cite{yang2020stochastic} also proposes stochastic primal-dual algorithm for solving AUC maximization with a general convex pairwise loss. They propose to use Bernstein polynomials \cite{powell1981approximation} to uniformly approximate a general loss. This reduction for AUC maximization with a general convex pairwise loss is equivalent to a weakly convex min-max problem (for learning a linear model). Then, the authors  apply the stochastic proximal point based method \cite{rafique2018non} for AUC maximization which has a per-iteration cost $\O(md)$, where $m$ is the degree of Bernstein polynomials used to approximate the original convex surrogate loss. Despite its non-convexity, they have proved its global convergence by exploring the appealing convexity-preserving property \cite{powell1981approximation} of Bernstein polynomials and the intrinsic structure of the min-max formulation. However, the final convergence in terms of the original objective function is of a slow rate $\O({1\over \sqrt{m}}).$

\subsection{Summary} Two main classes of methods have been proposed for stochastic AUC maximization: stochastic batch-pairwise (BP) methods and stochastic primal-dual (PD) methods. Stochastic batch-pairwise methods are generic which depend on the strategy of pairing examples while the stochastic PD methods explore the special problem structure which facilitates the design of fast stochastic optimization algorithms. We have compared different works in Table~\ref{table:stochastic} from different perspectives.

\section{Deep AUC Maximization (DAM): The Fourth Age}\label{sec:dam}
Recently, there is a surge of interest in AUC maximization for learning deep neural networks, i.e., deep AUC maximization (DAM). This problem has received much attention from the algorithmic perspective for solving the minimax objective of AUC maximization due to its advantage over the pairwise-loss based objective for big data. Then, it is employed for solving real-world classification problems (e.g., medical image classification) and achieves great success~\cite{DBLP:journals/corr/abs-2012-03173}.  Below, we will survey related works from algorithmic and practical perspectives. {We would like to point out that all algorithms surveyed below are also stochastic algorithms. However, the differences from works in the third age in that (i) algorithms presented below are applicable to any deep neural networks; in contrast, many algorithms in the third age are developed for learning  linear models by leveraging the special structure of the objective.; 
(ii) deep AUC maximization has faced some unique challenges, e.g., feature learning, regularization and normalization, etc., which will be discussed in Section~\ref{sec:out}.  }

 
\subsection{Non-Convex Concave Min-Max Optimization}\label{sec:minmax}
For deep learning, the prediction function $f_\w(\x)$ is a non-linear function of the model parameter $\w$, which makes the objective in~(\ref{eq:auc-emp}) and the minimax objective in~(\ref{opt:spp}) and ~(\ref{opt:spp2}) non-convex. Although standard stochastic methods (e.g., SGD, Adam) can be directly applied for solving the pairwise-loss based objective in~(\ref{eq:auc-emp}) with provable convergence to a stationary point, these methods are not directly applicable to the minimax objective in~(\ref{opt:spp}), which is more suitable for online learning and distributed optimization. The minimax objective~(\ref{opt:spp}) and~(\ref{opt:spp2}) is a non-convex strongly concave problem. Below, we will focus on stochastic methods for solving non-convex min-max problems, and we categorize different stochastic methods into two classes, i.e., two-loop proximal point based methods, and single-loop stochastic primal-dual methods. Without loss of generality, we consider the following min-max optimization problem for discussion: 
\begin{align}
    \min_{\w} F(\w):=\{\max_{\alpha\in\Omega}F(\w, \alpha) = \EX_{\z}[F(\w, \alpha; \z)]\}.
\end{align}

\begin{algorithm}[t]
\caption {A Unified Framework for solving $\min_{\w\in\R^d}\max_{\alpha\in\Omega}F(\w, \alpha)$} \label{alg:prox}
\begin{algorithmic}[1]
\STATE{Initialization: $\w_0\in \R^d, \alpha_0 \in \Omega, \gamma, T_1, \eta_1$,}
\FOR{$k=1,2, ..., K$}
\STATE{Set $\w_0^k = \w_{k-1}$, $\alpha_0^k = \Lambda(\w_{k-1},\alpha_{k-1})$\hfill{$\diamond \Lambda$ is a certain function}} 
\STATE{Construct $F_k(\w, \alpha) = F (\w, \alpha) + \frac{\gamma}{2} \|\w - \w_0^k\|^2$}
\STATE{Solve $(\w_k, \alpha_k)$ = $\mathcal A(F_k, \w^k_0, \alpha^k_0, \eta_k, T_k)$,\hfill{$\diamond \mathcal A$ is a stochastic algorithm}} 
\STATE{Decrease $\eta_k$ appropriately, increase $T_k$ accordingly}
\ENDFOR 
\STATE{Return $\w_K$.}
\end{algorithmic}
\label{alg:main}
\end{algorithm}

\paragraph{\bf Proximal Point Based Methods.} The proximal point based methods follow a common framework as shown in Algorithm~\ref{alg:prox}. This general framework has several unique features: (i) the algorithm is run in multiple stages $k=1, \ldots, K$; (ii) at each stage a quadratic regularized function $F_k(\w, \alpha)$ is constructed by adding a quadratic function $\frac{\gamma}{2}\|\w - \w_0^k\|^2$, where $\gamma>0$ is a proper hyperparameter; (iii) a proper stochastic algorithm $\mathcal A$ is employed for solving the regularized function with a step size $\eta_k$ and a number of iterations specified by $T_k$, whose output denoted by $\w_k, \alpha_k$ that are usually the last or the averaged solutions across all iterations in this stage; (iv) the step size $\eta_k$ and the number of iterations $T_k$ are changed appropriately for next stage. The following different methods differ in how to change $\eta_k, T_k$ and how to implement the function $\Lambda$ for computing $\alpha_0^k$.

Rafique et al.~\cite{rafique2018non} are the first to study non-convex concave min-max optimization problems and to establish the convergence rate. In particular, they assume the objective function $F(\w, \alpha)$ is weakly convex in terms of the primal variable $\w$ and is (strongly) concave in terms of the dual variable $\alpha$. A function is called weakly convex if it becomes a convex function by adding a quadratic function in term of the decision variable with a proper scaling factor. This is the motivation of adding $\frac{\gamma}{2}\|\w - \w_0^k\|^2$ to the objective at each stage, which can make the objective convex or strongly convex with an appropriate $\gamma>0$.   Since the objective function is non-convex and not  necessarily smooth,  they consider a convergence measure for weakly convex function, i.e., nearly stationary solution~\cite{doi:10.1137/18M1178244}. An $\epsilon$-level nearly stationary solution to a problem $\min_{\w}F(\w)$ is defined as a point $\w$ such that there exists a point $\widehat\w$ satisfying $\|\w - \widehat\w\|\leq O(\epsilon)$ and $\text{Dist}(0, \partial F(\widehat\w))\leq \epsilon$, where $\text{Dist}(\cdot,\cdot)$ denotes the Euclidean distance from a point to a set. In this work, the authors consider both the online setting and the offline (a.k.a. finite-sum) setting for the objective function $F(\w, \alpha)$. For the online setting, they employ stochastic mirror descent (SMD) method for implementing $\mathcal A$. The parameters are set as $\eta_k\propto  1/\sqrt{k}, T_k\propto k^2$ when the objective is only concave in terms of the dual variable $\alpha$, and are set as $\eta_k\propto  1/k, T_k\propto k$ when the objective is strongly concave in terms of the dual variable. When the objective is weakly convex and concave, the sample complexity is in the order of $O(1/\epsilon^6)$ for finding an $\epsilon$-level nearly stationary solution to $F(\w) = \max_{\alpha\in\Omega}F(\w, \alpha)$, and when the objective is strongly concave, they improve the sample complexity to $O(1/\epsilon^4 + C/\epsilon^2)$ by considering a special class such that $\alpha_*=\arg\max_{\alpha\in\Omega}F(\w, \alpha)$ can be computed, where $C$ denotes the complexity for computing $\alpha_*$ given $\w$. To enjoy this improved complexity, they compute $\alpha_0^k$ by solving $\max_{\alpha\in\Omega}F(\w_0^k, \alpha)$ to the optimal solution for $\alpha$. For the finite-sum setting with $n$ components for the function $F(\w, \alpha)$, they improve the complexity to $O(n/\epsilon^2)$ when the objective is strongly concave in terms of $\alpha$.

Yan et al.~\cite{yan2020sharp} further improve the algorithm and complexity for solving weakly convex and strongly concave min-max problems. They do not assume certain structure of the objective function or the optimal dual variable  can be easily computed given $\w$. Their algorithm is similar to the first algorithm proposed in~\cite{rafique2018non}, i.e., the initial solution $\alpha^k_0$ is simply the averaged solution from last stage of running $\mathcal A$, i.e., $\Lambda(\w_{k-1}, \alpha_{k-1})=\alpha_{k-1}$. They develop a novel analysis to prove the algorithm enjoys a sample complexity of $O(1/\epsilon^4)$ for finding an $\epsilon$-level nearly stationary solution to $F(\w)$. The key challenge lies at tackling error $\|\alpha^k_0 - \alpha(\w_k)\|^2$ in the upper bound for solving $\min_{\w}\max_{\alpha}F_k(\w,\alpha)$, where $\alpha(\w_k)= \arg\max_{\alpha\in\Omega}F(\w, \alpha)$. In~\cite{rafique2018non}, the authors compute $\alpha^k_0=\alpha(\w_0^k)$, which reduce the dual error $\|\alpha^k_0 - \alpha(\w_k)\|^2$ to $\|\w_{k-1} - \w_k\|^2$ due to the Lipchitz continuity of $\alpha(\w)$, which is decreasing to zero. In contrast, the authors of \cite{yan2020sharp} avoid computing $\alpha_0^k=\alpha(\w_0^k)$ instead directly set $\alpha_0^k=\alpha_{k-1}$. As a result, they need to explicitly tackle the error $\|\alpha^k_0 - \alpha(\w_k)\|^2$. To this end, they develop a novel analysis based on a new Lyapunov function to prove the convergence. In contrast to that in~\cite{rafique2018non} which uses the recursion of $F(\w_{k-1}) - F(\w_k)$, Yan et al. use both the recursions of the duality gap of the regularized function $F_k$ and of $F(\w_{k-1}) - F(\w_k)$. They are able to bound $\|\alpha^k_0 - \alpha(\w_k)\|^2$ by the duality gap of the regularized function.  

When the objective is just concave in terms of the dual variable, Zhao~\cite{zhao2020primal} develop a stagewise stochastic algorithm similar to Algorithm 1 except that the primal function is also smoothed by adding a strongly concave term on the dual variable, which has the same complexity as~\cite{rafique2018non}. 

Liu et al.~\cite{liu2019stochastic} consider the deep AUC maximization explicitly and develope the first practical and provable stochastic algorithms for deep AUC maximization based on the min-max formulation of the pairwise square loss function, which enjoy a faster convergence rate. In particular, they assume that the primal objective function $F(\w)$ satisfies a PL condition, i.e., there exists $\mu>0$ such that $\|\nabla F(\w)\|^2 \geq \mu (F(\w) - F_*)$, where $F_*$ denotes the global minimum  of $F$. They show that two-layers neural network satisfy this PL condition. Based on this condition, they have shown that Algorithm~\ref{alg:prox} enjoys a faster convergence rate in the order of $O(1/(\mu^2\epsilon))$ for finding an $\epsilon$-level optimal solution. For $\alpha_0^k$, they compute it similarly to that in~\cite{rafique2018non} except that it is approximated by sampling a number of data. For the parameters $\eta_k, T_k$, they decrease $\eta_k$ geometrically and increase $T_k$ geometrically. For the stochastic algorithm $\mathcal A$, they employ both stochastic primal-dual gradient method and stochastic primal-dual adaptive gradient method, where the latter one could enjoy even faster convergence when the stochastic gradients have a slow growth.  

Recently, Guo et al.~\cite{DBLP:journals/corr/abs-2006-06889} propose a family of  Proximal Epoch Stochastic (PES) methods for more generic non-convex min-max optimization under a PL condition and establish several improved rates under different conditions, e.g., near convexity condition of the primal objective, and Lipchitz condition of stochastic gradients. Under these conditions, they can reduce the sample complexity to $O(1/(\mu\epsilon))$. They also analyze the convergence rates for multiple stochastic algorithms $\mathcal A$, including stochastic gradient descent ascent, stochastic optimistic gradient descent ascent, stochastic primal-dual STORM updates, etc. In addition, they also establish the PL condition of the primal objective for AUC maximization for learning over-parameterized neural networks.

Guo et al.~\cite{DBLP:conf/icml/GuoLYSLY20,DBLP:conf/icml/YuanGXYY21} also study the federated deep AUC maximization by solving the min-max formulations in a distributed fashion, and establish both computation and communication complexity under a PL condition of the objective function. It is notable that \cite{DBLP:conf/icml/YuanGXYY21} claims that they achieve the optimal communication complexity.

\paragraph{\bf Single-loop Stochastic Primal-Dual Methods.} A generic framework of single-loop stochastic gradient descent ascent methods is shown in Algorithm~\ref{alg:single}. At each iteration, it computes a stochastic gradient estimator $\mathbf u_{t+1}$ of $\nabla_\w F(\w_t, \alpha_t)$ and then update the primal variable based on this gradient estimator. Then it computes a stochastic gradient estimator $\v_{t+1}$ of $\nabla_\alpha F(\widehat\w_t, \alpha_t)$ and update the dual variable based on $\mathbf v_{t+1}$ for some $\widehat\w_t$. Different methods differ from each other on how to compute the gradient estimators $\mathbf u_{t+1}$ and $\mathbf v_{t+1}$. 

\begin{algorithm}[t]
\caption {A Single-loop Algorithmic Framework for solving $\min_{\w\in\R^d}\max_{\alpha\in\Omega}F(\w, \alpha)$} \label{alg:single}
\begin{algorithmic}[1]
\STATE{Initialization: $\w_0\in \R^d, \alpha_0 \in \Omega, \eta_1, \eta_2, T$,}
\FOR{$t=0, 1,2, ..., T$}
\STATE Compute a stochastic estimator of $\nabla_\w F(\w_t, \alpha_t)$  by $\mathbf u_{t+1}$
\STATE Update $\w_{t+1} = \w_t - \eta_{1,t} \mathbf u_{t+1}$
\STATE Set $\widehat\w_t$ appropriately
\STATE Compute  a stochastic estimator of $\nabla_\alpha F(\widehat\w_t, \alpha_t)$ by $\mathbf v_{t+1}$ 
\STATE Update $\alpha_{t+1} = \prod_{\Omega}[\alpha_t - \eta_{2,t} \mathbf v_{t+1}]$
\ENDFOR 
\end{algorithmic}
\label{alg:main}
\end{algorithm}
Lin et al.~\cite{lin2019gradient} are the first to analyze the single-loop primal-dual method (the basic stochastic gradient descent ascent method, i.e., SGDA) for non-convex concave min-max optimization problems, corresponding to Algorithm~\ref{alg:single} with $\widehat\w_t=\w_t$. In the paper, they assume the objective function $F(\w, \alpha)$ is smooth in terms of both $\w$ and $\alpha$. They compute $\u_{t+1} = \frac{1}{B}\sum_{\z_t\in\mathcal B}\nabla_\w F(\w_t, \alpha_t, \z_t)$ and $\v_{t+1} =\frac{1}{B}\sum_{\z_t\in\mathcal B } \nabla_\alpha F(\w_t, \alpha_t, \z_t)$ based on  a batch of $B$ samples. However, their convergence results are un-satisfactory. In particular, for non-convex concave min-max problems, their analysis yields an $O(1/\epsilon^8)$ complexity for finding an $\epsilon$-stationary solution to $F(\w)$; and for non-convex strongly concave min-max problems, their analysis requires a large mini-batch size in the order of $O(1/\epsilon^2)$ and yields an $O(1/\epsilon^4)$ sample complexity.  It is worth to point out that the complexity for the former case is worse than that established in~\cite{rafique2018non} and the complexity for the latter case matches that in~\cite{rafique2018non} but requires a large mini-batch size, which is not required in~\cite{rafique2018non}.  Recently, Bo{\c{t}} and B{\"o}hm~\cite{2020arXiv200713605I} extend the analysis to stochastic alternating (proximal) gradient descent ascent method which uses $\widehat\w_t = \w_{t+1}$ to compute the estimator $\v_{t+1}$. However, this algorithm suffers from the same issue of requiring a large mini-batch size and the worse complexity for non-convex concave min-max problems.

Recently, Guo et al.~\cite{guo2021stochastic} develop a new stochastic primal-dual method for solving non-convex strongly concave min-max problems under the smoothness assumption of $F(\w, \alpha)$. They address the issue of large mini-batch size requirement in~\cite{lin2019gradient,2020arXiv200713605I}. The key improvement lies at using moving average to compute the estimator $\u_{t+1}$, i.e., $\u_{t+1} = (1-\beta_{1,t})\u_t + \beta_{1,t} \O_\w(\w_t, \alpha_t)$, and simply 
use $\v_{t+1} = \O_\alpha (\w_t, \alpha_t; \z_t)$, where $\O_\w$ and $\O_\alpha$ denote an unbiased stochastic estimator of $\nabla_\w F(\w, \alpha)$ and $\nabla_\alpha F(\w, \alpha)$, respectively.  The authors also establish the convergence using adaptive step sizes such as the Adam-style with a sample complexity in the order of $O(1/\epsilon^4)$. This is the first work that establishes the convergence Adam-style updates for solving non-convex min-max problems. 

An improved complexity of $O(1/\epsilon^3)$ is achieved in several recent works under the Lipschitz continuous assumption for the stochastic gradient $\nabla_\w F(\w, \alpha; \z)$ and $\nabla_\alpha F(\w, \alpha; \z)$~\citep{luo2020stochastic,DBLP:journals/corr/abs-2008-08170}, which is a stronger condition than the smoothness condition of the objective function. Luo et al.~\cite{luo2020stochastic} are the first to establish such an improved rate. Their algorithm called SREDA uses the SPIDER/SARAH technique~\cite{fang2018spider,nguyen2017sarah} to update the gradient estimators $\u_{t+1}$ and $\v_{t+1}$, i.e., $\u_{t+1} = \u_t + \frac{1}{B}\sum_{\z_t\in\mathcal B}\nabla_\w F(\w_t, \alpha_t, \z_t) - \frac{1}{B}\sum_{\z_t\in\mathcal B}\nabla_\w F(\w_{t-1}, \alpha_{t-1}, \z_t)$, where $B$ is in the order of $O(1/\epsilon)$. $\u_t$ and $\v_t$ are re-computed based on a large batch size in the order of $O(1/\epsilon^2)$ every $q=O(1/\epsilon)$ iterations. It is worth mentioning that SREDA is a double loop algorithm, where the inner loop is to mainly update the dual variable and the estimators $\u_{t+1}, \v_{t+1}$ with multiple iterations and the outer loop is to update the primal variable.  This issue was addressed by Huang et al.~\cite{DBLP:journals/corr/abs-2008-08170}, who propose a single-loop algorithm named AccMDA to  enjoy a fast rate of $O(1/\epsilon^3)$ under the Lipschitz continuous assumption for the stochastic gradient. They use the STORM technique~\cite{cutkosky2019momentum} to compute $\u_{t+1}, \v_{t+1}$, i.e., $\u_{t+1} = (1-\beta_t)\u_t + \beta\nabla_\w F(\w_t, \alpha_t, \z_t) - \beta_t\nabla_\w F(\w_{t-1}, \alpha_{t-1}, \z_t)$, similarly for $\v_{t+1}$. It is notable that both SREDA and AccMDA require computing two (batch) stochastic gradients at each iteration. It is notable that AccMDA has a worse dependence on the strong concavity parameter than that in~\cite{guo2021stochastic,lin2019gradient}. It is likely that by simply computing $\v_{t+1}=\nabla_\alpha F(\w_t, \alpha_t, \z_t)$ in AccMDA, one should be able to improve the dependence on the strong concavity as in~\cite{guo2021stochastic}.

Yang et al.~\cite{yang2020global} develop a single-loop algorithm for improving the convergence rate of non-convex min-max optimization under PL conditions. They consider a class of smooth non-convex non-concave problems, which satisfy both the dual-side PL condition (i.e., $F(\w, \cdot)$ satisfies a PL condition for any $\w$) and the primal-side  PL condition (i.e., $F(\cdot, \alpha)$ satisfies a PL condition for any $\alpha$).  They propose stochastic alternating gradient descent ascent algorithm (Stoc-AGDA)  and establish a global convergence for a Lyapunov function $F(\w_t) - F_* + \lambda (F(\w_t) - F(\w_t, \alpha_t))$ for a constant $\lambda$ in the order of $O(1/\epsilon)$, which directly implies the convergence for the primal objective gap in the same order. Their algorithm uses a polynomially decreasing or very small step sizes. It is notable that the complexity of Stoc-AGDA is worse than that of PES established in~\cite{DBLP:journals/corr/abs-2006-06889} under similar PL conditions but requiring the strong concavity of the objective function in terms of the dual variable, which makes  PES more appropriate to deep AUC maximization. Without the primal-side PL condition, Stoch-AGDA and Smoothed-AGDA are also analyzed under the dual-side PL condition with a better dependence on the condition number~\cite{DBLP:journals/corr/abs-2112-05604}. 


\paragraph{\bf Improved Rates for the Offline (Finite-sum) Setting.} There are also multiple papers  trying to improve the complexity of non-convex (strongly) concave min-max optimization in the finite-sum setting by leveraging the variance reduction techniques~\cite{rafique2018non,luo2020stochastic,yang2020global}. However, they usually require computing the gradient based on the full-batch or a large-batch that are less practical for deep learning with big data.

\begin{table*}[t] 
	\caption{Comparison of different stochastic methods for solving non-convex strongly concave min-max optimization. $\epsilon$ denotes the target accuracy level for the objective gradient norm, i.e., $\EX[\|\nabla F(\w)\|]\leq \epsilon$ or the primal objective gap, i.e., $\EX[F(\w) -  F(\w_*)]\leq \epsilon$. The column ``oracle" denotes the condition on the stochastic gradient. $\mu$ denotes a parameter in the assumed PL condition. 
}\label{tab:0} 
\vspace*{-0.1in}
	\centering
	\label{tab:1}
	\scalebox{0.85}{\begin{tabular}{llllccc}
			\toprule
	        Method&Category	&batch size&\makecell{Sample\\ Complexity} &Oracle&\makecell{Experiments for\\ AUC Max.}\\
		    \hline 
			SGDA~\citep{lin2019gradient}&Single-loop&$O(1/\epsilon^2)$&$O(1/\epsilon^4)$&General&No\\
			PDAda~\cite{guo2021stochastic}&Single-loop&$O(1)$&$O(1/\epsilon^4)$&General & Yes\\
			AccMDA~\citep{DBLP:journals/corr/abs-2008-08170}&Single-loop&$O(1)$&$O(1/\epsilon^3)$&Lipschitz&No\\
			Stoc-AGDA~\cite{yang2020global}&Single-loop&$O(1)$&$O(1/(\mu^2\epsilon))$&General&No\\
			\midrule
		    \midrule
		    PG-SMD~\cite{rafique2018non}&ProximalPoint&$O(1)$&$O(1/\epsilon^4+n/\epsilon^2)$&General&No\\
		    Epoch-SGDA~\citep{yan2020sharp}&Proximal Point&$O(1)$&$O(1/\epsilon^4)$&General&No\\
		    PPD-SG~\cite{liu2019stochastic}&Proximal Point&$O(1)$&$O(1/\epsilon^4)$&General & Yes\\
		    PPD-AdaGrad~\cite{liu2019stochastic}&Proximal Point&$O(1)$&$O(1/\epsilon^4)$&General & Yes\\
		    PES-$\mathcal A$~\cite{DBLP:journals/corr/abs-2006-06889}&Proximal Point&$O(1)$&$O(1/\epsilon^4)$$\sim$$O(1/(\mu\epsilon))$&\makecell{General\\or Lipchitz} & Yes\\
			\bottomrule
	\end{tabular}}
\end{table*}

\subsection{Deep Partial AUC Maximization}
Deep pAUC maximization is challenging not only because of the non-differentiable selection operator but also due to non-convexity of the objective. 
Below, we discuss two classes of methods.  

\paragraph{\bf Naive Mini-batch Approach.} \citet{10.5555/2968826.2968904} propose mini-batch based stochastic methods for pAUC maximization, which is applicable to deep learning. At each iteration, a gradient estimator is simply computed based on the pAUC surrogate function of the mini-batch data. However, this heuristic approach is not guaranteed to converge for minimizing the pAUC objective and its error scales as $O(1/\sqrt{B})$, where $B$ is the mini-batch size.  Ueda and Fujino~\cite{Ueda2018PartialAM} consider partial AUC maximization for learning non-linear scoring functions, e.g., neural networks and probabilistic generative models. The paper claims to use the Adam optimizer~\cite{kingma2014adam} in Tensorflow for optimizing the partial AUC. However, it does not provide any discussion how the algorithm was implemented and what is the complexity and convergence of the optimization algorithm. We conjecture they use the naive mini-batch approach equipped with the Adam optimizer. For experiments, they have used an image dataset namely Hyper Suprime-Cam (HSC) dataset~\cite{Morii2016MachinelearningSO}  with 487 real  and 267,074 bogus  optical transient objects collected with the HSC using the Subaru telescope. 

\paragraph{\bf Reduction Approaches.} The idea  is to reduce the objective into different formulations (equivalent or approximate), which facilitate the design of large-scale optimization algorithms. 

Recently, Yang et al. (2021) \cite{pmlr-v139-yang21k} consider optimizing two-way pAUC with FPR less than $\beta$ and TPR larger than $1-\alpha$. The paper focuses on simplifying the optimization problem that involves selection of top ranked negative examples and bottom ranked positive examples. They first formulate the problem into a bilevel optimization, where the upper level objective function is a weighted average of pairwise surrogate loss and the lower level optimization problem is to compute the weights that accounts for selection of top ranked negative examples and bottom ranked positive examples. To address the computational challenge for solving the bilevel optimization problem, the authors propose to simplify the lower level problem by relaxing the non-decoposable constraint on the decision variables into decomposable regularization. As a result, a simplified weighted pairwise loss minimization problem is derived, where the weights for each positive-negative pair is a product of two individual weights  that are computed directly from the prediction scores of the positive and negative examples using a penalty function. Then any stochastic algorithms based on random positive-negative pairs can be employed for solving their formulation, e.g.,  SGD, Adam.


Zhu et al.~\cite{otpaUC} consider both pAUC maximization and two-way pAUC maximization. For pAUC maximization, they focus on that with FPR in a range $(0, \beta)$. They propose two formulations for pAUC maximization by leveraging distributionally robust optimization technique, and develope stochastic algorithms for optimizing both formulations for both  pAUC and two-way pAUC. In particular, for pAUC they define  a robust loss for each positive data by 
\begin{align*}
&\hat L_\phi(\w; \x_i)=\max_{\p\in\Delta_{n_-}} \sum_{\x_j\in\S_-} p_j\ell(f_\w(\x_j) - f_\w(\x_i)) - \lambda D_\phi(\p, 1/n_-),
\end{align*}
where $\Delta_{n_-}=\{\p\in\R^{n_-}: \sum_j p_j =1, p_j\geq 0\}$ is a simplex, $D_\phi(\p, 1/(n_-)) = \frac{1}{n_-}\sum_i \phi(n_-p_i)$ is a divergence measure defined by a function $\phi$.  Then the following objective is used  for one-way pAUC maximization:
\begin{align}\label{eqn:pauces}
\min_{\w}\frac{1}{n_+}\sum_{\x_i\in\S_+}\hat L_\phi(\w; \x_i).
\end{align}
They consider two functions $\phi$, i.e.,  the KL divergence $\phi_{kl}(t) = t\log t - t + 1$, which gives $D_\phi(\p, 1/n_-) =\sum_ip_i\log (n_-p_i)$,  and the CVaR divergence $\phi_{c}(t) =\mathbb I(0<t\leq 1/\beta)$ with a parameter $\beta\in(0,1)$, which gives $D_\phi(\p, 1/n_-) = 0$ if $p_i\leq 1/(n_-\beta)$ and infinity otherwise. It is shown that if $\ell(\cdot)$ is non-decreasing, when using $\phi_{c}$ the objective~(\ref{eqn:pauces}) is equivalent to~(\ref{eq:pauc-emp}) for pAUC maximization, when using $\phi_{kl}$ it gives a soft estimator of pAUC. 

For solving~(\ref{eqn:pauces})  with CVaR divergence, they formulate the problem as a weakly convex optimization problem by introducing another set of variables, i.e., 
\begin{align}\label{eqn:opauccvar} 
\hspace*{-0.1in}\min_{\w}\min_{\mathbf s\in\R^{n_+}} F(\w, \mathbf s)= \frac{1}{n_+}\sum_{\x_i\in\S_+} \left(s_i  +  \frac{1}{\beta}  \frac{1}{n_-} \sum_{\x_j\in \S_-}(\ell(f_\w(\x_j) - f_\w(\x_i)) - s_i)_+\right).
\end{align}
They develop an efficient stochastic algorithm named SOPA with a sample complexity of $O(1/\epsilon^4)$ for finding a nearly $\epsilon$-stationary point for $F(\w, \mathbf s)$. 

For solving~(\ref{eqn:pauces})  with KL divergence, they formulate the problem as a novel finite-sum coupled compositional optimization  problem, i.e., 
\begin{align}\label{eqn:pauceskl}
\min_{\w}\frac{1}{n_+}\sum_{\x_i\sim\S_+}\lambda\log \EX_{\x_j\in\S_-}\exp(\frac{\ell(f_\w(\x_j) - f_\w(\x_i))}{\lambda}).
\end{align}
A stochastic algorithm named SOPA-s is proposed for solving~(\ref{eqn:pauceskl}) with  a sample complexity of $O(1/\epsilon^4)$ for finding an  $\epsilon$-level stationary point. 

For two-way pAUC such that FPR is less than $\beta$ and TPR is larger than $\alpha$, the authors further define a new objective: 
\begin{align}\label{eqn:tpaucdro}
F(\w; \phi, \phi')=\max_{\mathbf q\in\Delta_{n_+}}\sum_{\x_i\in\S_+} q_i \hat L_\phi(\x_i, \w)  -  \lambda' D_{\phi'}(\mathbf q, \frac{1}{n_+}).
\end{align}
They prove that when $\phi(t)=\mathbb I(0<t\leq 1/\beta)$ and $\phi'(t)=\mathbb I(0<t\leq 1/\alpha)$ the above objective is equivalent to~(\ref{eq:tpauc-emp}) if $\ell(\cdot)$ is non-decreasing. The authors develop two algorithms for solving the above objective with CVaR divergence and KL divergence, respectively, and establish their convergence. A sample complexity of $O(1/\epsilon^4)$ is established for the algorithm that optimizes the above objective with the KL-divergence in order to find a $\epsilon$-level stationary point. For optimizing the above objective with CVaR divergence, the sample complexity of their algorithm is $O(1/\epsilon^6)$. 
This is the first time that stochastic algorithms are developed for optimizing two-way pAUC for deep learning with convergence guarantee. 

A concurrent work by Yao et al.~\cite{paUCyao} focuses on optimizing pAUC such that FPR is in a range $(\alpha, \beta)$. When $\ell(\cdot)$ is a non-decreasing function,  they formulate the problem~(\ref{eqn:pauces}) as non-smooth different-of-convex problems: 
\begin{align}\label{eqn:pAUCr}
    F(\w) = \frac{1}{n_+}\sum_{\x_i\in\S_+} (F(\w; \x_i, k_2) - F(\w; \x_i, k_1)),
\end{align}
where 
\begin{align*}
   F(\w; \x_i, k)= \frac{1}{n_-}\sum_{\x_j\in\S^\downarrow_-[1, k]}\ell(f_\bw (\x_j) - f_\bw(\x_i)) = \min_{\lambda} \frac{\lambda k}{n_-} +  \frac{1}{n_-}\sum_{\x_j\in\S_-} (\ell(f_\bw (\x_j) - f_\bw(\x_i)) -\lambda)_+.
\end{align*}
They develop an efficient approximated gradient descent method based on the Moreau envelope smoothing technique, inspired by recent advances in non-smooth DC optimization~\cite{sun2021algorithms}. To increase the efficiency of large data processing, they use an efficient stochastic block coordinate update for solving each sub-problem inexactly. A sample complexity of $O(1/\epsilon^6)$ is established for their algorithm in order to find a nearly $\epsilon$-stationary solution.

\subsection{Applications of Deep AUC Maximization (DAM)}
Due to the success of deep learning in various applications, DAM has also been applied to different domains with demonstrated success. Below, we review some applications of DAM.


\paragraph{\bf Medical Image Classification.} \citet{sulam2017maximizing} consider the classification of breast cancer based on imbalanced mammogram images. They learn a deep convolutional neural network by AUC maximization by using the online buffered gradient method proposed by Zhao et al~\cite{Zhaoicml11}.  Nevertheless, the issue of this approach is that it cannot scale to large datasets as it requires a large buffer to store positive and negative samples at each iteration for computing an approximate AUC score. As a result, they only consider small-scale datasets. In particular,  two datasets are used.  The first one named IMG is a proprietary mammogram dataset comprising of 796 patients, 80 of them defined as positive (164 images), and 716 negative (1869 images) with both Cranial-Caudal (CC) and Mediolateral-Oblique (MLO) views, belonging to normal patients as well as benign findings. The second dataset is the public INbreast dataset, which consists of 115 cases with 410 images.

Yuan et al.~\cite{DBLP:journals/corr/abs-2012-03173} are the first to evaluate the performance of DAM on large-scale medical image data with hundreds of thousands of images for learning modern deep neural networks (e.g., ResNets, DenseNets). They propose a new minimax objective as in~(\ref{opt:spp2}) for robust AUC maximization to alleviate the issues of the square loss, namely the sensitivity to noisy data and the adverse effect on easy data. The new objective is shown to be more robust than the commonly used  square loss, while enjoying the same advantage in terms of large-scale stochastic optimization. The authors employ PESG~\cite{DBLP:journals/corr/abs-2006-06889} for solving the minimax objective.  They conduct extensive empirical studies of DAM on four difficult medical image classification tasks discussed below.  
\begin{itemize}
\vspace*{-0.05in}\item CheXpert Competition.  {CheXpert} is a large-scale chest X-ray dataset for detecting chest and lung diseases, which is released through a medical AI competition~\cite{irvin2019chexpert}. The training data consists of 224,316 high-quality X-ray images from 65,240 patients with frontal and lateral views. The validation dataset consists of 234 images from 200 patients. The testing data has images for 500 patients, which is not released to the public. The model is only evaluated for predicting 5 selected diseases, i.e., Cardiomegaly, Edema, Consolidation, Atelectasis, Pleural Effusion, which have an average imbalance ratio (i.e., the proportion of positive examples) of 20.21\% in the training set. AUC and NRBC are used for evaluation, where NRBC refers a number of radiologists out of 3 are beaten by AI algorithms. Yuan et al.~\cite{DBLP:journals/corr/abs-2012-03173} achieved 1st place using their DAM method in this competition in August 2020. Compared with the standard deep learning approach that minimizes the cross-entropy loss, their DAM method achieves 2\% improvements. 
\item  SIIM-ISIC Melanoma Classification.  {Melanoma} is a skin cancer and is the major cause of skin cancer death~\cite{miller2006melanoma}. Kaggle hold a  competition for melanoma classification in 2020.   The dataset consists of 33,126 training images with 584 malignant melanoma images and 10,892 testing images with an unknown number of malignant  melanoma images. The testing set is split into a validation set with 30\% images  and the final testing set with 70\% images. The raw images have high resolutions, e.g., 6000x4000.  Yuan et al.~\cite{DBLP:journals/corr/abs-2012-03173} demonstrate the performance DAM for Melanoma Classification.  They resize the the image into lower resolutions, e.g., 384x384 and also use additional 12,859 images from previous competitions in their experiments. Their method achieves the 33rd place out 3314 teams for the competition by ensemble over 10 models. A simple ensemble of a DAM model and a standard model learned by optimizing the cross-entropy loss beats the winning team by combining 18 models~\cite{DBLP:journals/corr/abs-2012-03173}. Compared with the standard deep learning approach that minimizes the cross-entropy loss, the DAM method achieves 1\% improvements. 

\item Breast-Cancer Screening. For this task, they use the DDSM+ data, which is a combination of two datasets namely DDSM and CBIS-DDSM \cite{bowyer1996digital,heath1998current}. The dataset consists of 55,000 mammographic images (224$\times$224) taken at lower doses than usual X-rays for training with an imbalance ratio of 13\% and 13,900 images for testing with an imbalance ratio of 4\%. Compared with the standard deep learning approach that minimizes the cross-entropy loss, the DAM method achieves 1.5\% improvements. 

\item Lymph Node Tumor Detection. They use the PathCamelyon dataset for this task, which consists of 294,912 color images (96$\times$96) extracted from histopathologic scans of lymph node section for training and 32,768 images for testing with balanced class ratio~\cite{veeling2018rotation, bejnordi2017diagnostic}. The authors manually construct an imbalanced dataset with an imbalance ratio of 1\% for their experiments. Compared with the standard deep learning approach that minimizes the cross-entropy loss, the DAM method achieves 5\% improvements. 
\end{itemize}

Recently, \cite{he2021performance} investigates DAM for {COVID-19 Chest X-ray Classification}. Covid-19 is a global pandemic that broke out in 2020. Early detection of Covid-19 is crucial to contain the spread of the virus and is helpful for providing early treatment for the patients with Covid-19. The authors use a  dataset (COVIDx8B), which consists of 15,952 chest X-ray images for training with 13.5\% Covid-19 positive samples and 400 images for testing with balanced positive and negative samples. The authors use self-supervised training method discussed below for learning a backbone network and then use the LibAUC library~\cite{DBLP:journals/corr/abs-2012-03173} for finetuning the network with significant improvements observed over the baseline  method.

\begin{table}[t]
    \centering
        \caption{{Benchmark Datasets for deep AUC and pAUC maximization. The differences between the MoleculeNet and OGBG molecular graph datasets lie at the way of training/testing split. The numbers in the last column are the best results on testing data reported in the referred papers for AUC maximization. We do not include the results for partial AUC maximization as it requires setting a threhold for FPR/TPR. We also do not present the results on datasets with manual construction of imbalanced data denoted by ``Artificial" for the ``Imbalance" column. }}
        \vspace*{-0.1in}
    \label{tab:benchmark}
    \resizebox{0.9\textwidth}{!}{
    \begin{tabular}{lllllll}
    \hline
       Dataset & Type&\makecell{Order of \\Data Size}& Imbalance&Source&\makecell{References for\\ DAM}&\makecell{Benchmark\\ Result}\\
        \hline
        STL10 &Natural Image& $10^3$  & Artificial & \cite{pmlr-v15-coates11a}& \cite{zhubenchmark,DBLP:journals/corr/abs-2012-03173} & -\\
        CIFAR10 &Natural Image& $10^4$  & Artificial&\cite{krizhevsky2009learning}  & \cite{zhubenchmark,DBLP:journals/corr/abs-2012-03173} & -\\
      CIFAR100 &Natural Image& $10^4$  & Artificial  &\cite{krizhevsky2009learning}& \cite{zhubenchmark,DBLP:journals/corr/abs-2012-03173}& -\\
        Cat vs Dog&Natural Image & $10^4$  & Artificial& \cite{asirra-a-captcha-that-exploits-interest-aligned-manual-image-categorization} & \cite{zhubenchmark,DBLP:journals/corr/abs-2012-03173} &-\\
        \hline
        Melanoma  &Skin Lesion Image&$10^4$  & Natural&\cite{kaggle}  & \cite{DBLP:journals/corr/abs-2012-03173,zhubenchmark} &0.9505 \\
        CheXpert  &Chest X-ray Image&$10^5$  & Natural&\cite{irvin2019chexpert} & \cite{DBLP:journals/corr/abs-2012-03173,zhubenchmark} &0.9305\\
        DDSM+&Mammographic Image &$10^4$& Natural&\cite{bowyer1996digital,heath1998current}& \cite{DBLP:journals/corr/abs-2012-03173}&0.9544\\ 
        PatchCamelyon&Microscopic Image & $10^5$& Artificial&\cite{veeling2018rotation, bejnordi2017diagnostic}&\cite{DBLP:journals/corr/abs-2012-03173}&- \\
        \hline
        MoleculeNet/HIV  &Molecular Graph&   $10^4$  & Natural &\cite{wu2018moleculenet}& \cite{zhubenchmark}&0.770\\
        MoleculeNet/PCBA &Molecular Graph&   $10^5$  & Natural &\cite{wu2018moleculenet}& -&-\\
        MoleculeNet/MUV &Molecular Graph &   $10^4$   & Natural &\cite{wu2018moleculenet}& \cite{zhubenchmark}&0.644\\
        MoleculeNet/Tox21  &Molecular Graph& $10^3$   & Natural &\cite{wu2018moleculenet}& \cite{zhubenchmark}&-\\
        MoleculeNet/ToxCast &Molecular Graph& $10^3$ & Natural  &\cite{wu2018moleculenet}& \cite{zhubenchmark}&-\\
        \hline
        ogbg-molhiv&Molecular Graph& $10^4$ & Natural  &\cite{hu2020ogb}& - &-\\
        ogbg-molpcba&Molecular Graph& $10^5$ & Natural  &\cite{hu2020ogb}& \cite{quanqimbmmo} &0.8406\\
        ogbg-molmuv &Molecular Graph &   $10^4$   & Natural &\cite{hu2020ogb}& \cite{otpaUC}&-\\
        ogbg-moltox21  &Molecular Graph& $10^3$   & Natural &\cite{hu2020ogb}& \cite{otpaUC}&-\\
        ogbg-moltoxcast &Molecular Graph& $10^3$ & Natural  &\cite{hu2020ogb}& \cite{otpaUC}&-\\
        \hline
    \end{tabular}}
\end{table}

\paragraph{\bf Molecular Property Predictions.}
Molecular property prediction is one of the key tasks in cheminformatics and has applications in many fields, including quantum mechanics, physical chemistry, biophysics and physiology.  Multiple molecular datasets have been released, e.g., MoleculeNet benchmark datasets (e.g., PCBA, HIV, MUV, Tox21, ToxCast)~\citep{wu2018moleculenet}, MIT AICURES Challenge dataset~\footnote{\url{https://www.aicures.mit.edu/tasks}},  Stanford OGB benchmark datasets (e.g., OGBG-molhiv, OGBG-molpcba)~\cite{DBLP:journals/corr/abs-2005-00687}. Recently,  \cite{wang2020moleculekit} has employed the LibAUC library~\cite{DBLP:journals/corr/abs-2012-03173} for solving the molecular property prediction and achieved {the 1st place  at MIT AICURES Challenge}.  Several research groups have also used LibAUC library for improving the performance on the OGBG-molhiv dataset~\footnote{\url{https://ogb.stanford.edu/docs/leader_graphprop/}}.   The authors of \cite{otpaUC} also consider pAUC maximization on some of these molecular datasets and compare different methods for pAUC maximization. 

{\paragraph{\bf Fraud/Outlier Detection} Identifying outliers in data is referred to as outlier or anomaly detection. It can be regarded as an extremely imbalanced classification problem where the object of interest (anomaly/outlier) is the minority class. AUC maximization can naturally be applied for outlier or anomaly detection. In \cite{ding2015adaptive}, multiple online AUC maximization algorithms including OAM \cite{Zhaoicml11} and OPAUC \cite{gao2013one}  are applied to the benchmark datasets for outlier/anomaly detection such as Webspam \cite{wang2012evolutionary}, Sensor Faults \cite{michaelides2009snap}, and Malware App \cite{zhou2012dissecting}. The performance of different stochastic AUC maximizaiton algorithms is compared in the studies \cite{natole2020fast,lei2021stochastic} for  anomaly detection. In a recent work \cite{huang2022auc}, the authors consider
AUC maximization for fraud detection on a graph. They consider learning both the parameters of a graph neural network (GNN) and a policy of edge pruning that affects prediction outputs of  the GNN. For learning the parameters of GNN, they employ the PPD-SG algorithm~\cite{liu2019stochastic} to solve the saddle point formulation. The learning of the edge pruner is formulated as a reinfocement learning problem and a classic policy gradient is used.  

\paragraph{\bf Other Applications.} Besides medical image classification, molecular property prediction, and fraud/outlier detection, AUC/pAUC maximization have been investigated in other applications, which do not necessarily involve deep learning. Examples include points of interest recommendation~\cite{8731461}, credit scoring for financial institutions~\cite{zhoucreditauc}, time series classification for medicine, manufacturing, and maintenance~\cite{ltspauc2020}, protein disorder prediction~\cite{10.1093/bioinformatics/btw446},   pedestrian detection \cite{paisitkriangkrai2013efficient}, differentiated gene detection \cite{liu2010partial}, and  discovery of motifs~\cite{10.1093/bioinformatics/btx255}.

\paragraph{\bf Benchmark and Library.} We present a summary of benchmark datasets in Table~\ref{tab:benchmark} for DAM and include references that provide benchmark results of deep AUC maximization and deep pAUC maximization. 
 The author T. Yang's group has developed an open-source library for DAM called LibAUC~\footnote{\url{www.libauc.org}}, which implements a set of efficient stochastic algorithms for deep AUC maximization, deep one-way pAUC maximization and deep two-way pAUC maximization.}

\subsection{Summary}
The research of deep AUC maximization springs from the studies for solving the non-convex min-max problems. The applicability in deep AUC maximization motivates a wave of studies for algorithmic design and theoretical analysis of  non-convex strongly concave min-max problems. New formulations for AUC  maximization and partial AUC maximization are also developed, for which efficient stochastic algorithms are proposed. The algorithms are then employed for solving real-world applications with great success, e.g., medical image classification and molecular property prediction.  However, there are still many challenges regarding deep (partial) AUC maximization to be addressed, which will be discussed in next section.

\section{Other Issues for DAM and Outlook for Future Work}\label{sec:out}
Below, we discuss five remaining or emerging  issues for DAM. 

\vspace*{-0.1in}
\paragraph{\bf Large-scale Stochastic  Optimization.} Although large-scale optimization algorithms for DAM have been developed, there are still many open problems to be addressed.  Below, we will list several important questions. (i) How to further improve the algorithms and theories for solving the composite objectives of AUC and for solving pAUC objectives? Zhu et al.~\cite{zhubenchmark} employ stochastic compositional algorithms for solving the composite objectives~(\ref{opt:obj3}). There are still much rooms for improving the optimization for solving pAUC objectives, e.g.,~(\ref{eqn:opauccvar})$\sim$(\ref{eqn:pAUCr}), or new formulations of pAUC.   (ii) How to optimize AUC in the federated learning setting? Although federated deep AUC maximization for the minimax objective has been considered in~\cite{DBLP:conf/icml/GuoLYSLY20,DBLP:conf/icml/YuanGXYY21},  federated learning algorithms for optimizing other objectives remains to be developed. In particular, the objectives (\ref{eqn:opauccvar})$\sim$(\ref{eqn:pAUCr}) for pAUC maximization are much more challenging to be optimized in the federated learning setting. 

\vspace*{-0.1in}
\paragraph{\bf Network Structures.} Standard deep neural networks have been used for DAM in different applications. For example, deep convolutional neural networks such as VGG, ResNets, DenseNets, EfficientNets have been used for medical image classification~\cite{sulam2017maximizing,DBLP:journals/corr/abs-2012-03173,he2021performance}. Graph neural networks e.g., graph isomorphism network (GIN) \cite{xu2018how,hu2019strategies}, Message-Passing Neural Network (MPNN)~\cite{wang2020moleculekit,pmlr-v70-gilmer17a}, have been used for molecular property prediction~\cite{otpaUC,wang2020moleculekit}. It remains to be explored by using more advanced network structures, e.g., vision transformer~\cite{dosovitskiy2021an} for medical image classification tasks in the context of DAM. Another interesting direction is to explore neural architecture search (NAS)~\cite{DBLP:journals/corr/abs-2005-11074} in the context of DAM. A natural question is if we use AUC as a performance measure of NAS, how would the found network be different from standard approaches that use accuracy as a performance measure. 

\vspace*{-0.1in}
\paragraph{\bf Regularization and Normalization.} Regularization is an important technique for improving generalization. A standard regularization technique is to use weight decay~\cite{GoodBengCour16}. It was shown to be effective for DAM as well~\cite{zhubenchmark}. Nevertheless, more algorithmic regularization techniques  should be considered for DAM, including explicit and implicit regularization. Recently, Zhu et al.~\cite{zhubenchmark} demonstrate that the explicit regularization by adding the quadratic term in the proximal point methods discussed in Section~\ref{sec:minmax} is helpful for improving the generalization. Implicit regularization by gradient based methods~\cite{DBLP:journals/corr/abs-1709-01953} for DAM is an interesting question to be explored. Normalization is another key technique for improving the training of deep neural networks, e.g., batch normalization~\cite{pmlr-v37-ioffe15}, layer normalization~\cite{Ba2016LayerN}, etc. Of particular interest to DAM is the normalization in the output layer. Yuan et al.~\cite{DBLP:journals/corr/abs-2012-03173} have used a batch score normalization layer, which normalizes the non-activated scores in the mini-batch such that that the $\ell_2$ norm of scores in the mini-batch is one. This is found to be better than not using any activation or normalization. Zhu et al.~\cite{zhubenchmark} show that using a sigmoid activation in the last layer also yields better performance than not using any activation or normalization. They also demonstrate that using the sigmoid activation is competitive with the batch score normalization, and they are better than the standard batch normalization.

\vspace*{-0.1in}
\paragraph{\bf Data Sampling and Augmentation.} The standard data sampling for deep learning is to use data shuffling over all examples. However, different data sampling strategies have been considered for imbalanced data, e.g., oversampling and undersampling~\cite{johnson2019survey}. Recently, Zhu et al.~\cite{zhubenchmark} demonstrate via empirical studies that oversampling for the minority class is helpful for improving the generalization performance of DAM. However, how can we incorporate more advanced oversampling or data augmentation techniques into DAM remains an interesting topic. One might consider synthetic oversampling method SMOTE~\cite{10.5555/1622407.1622416} and MIXUP~\cite{zhang2018mixup} for DAM. 

\vspace*{-0.1in}
\paragraph{\bf Feature Learning.} Feature learning is an important capability of deep learning for tackling un-structured  data. It is shown that directly optimizing the AUC loss from scratch does not necessarily yield better feature representations~\cite{yuan2022compositional}.  A practice of DAM uses a two-stage approach: the first stage is to learn the encoder network by optimizing the traditional cross-entropy loss and the second stage is to fine tune the encoder network and to learn the classifier by DAM~\cite{DBLP:journals/corr/abs-2012-03173}.  It is still not fully understood why optimizing the AUC loss in an end-to-end fashion does not yield better feature representations, and it remains an open problem how to learn better encoder networks by using DAM. Recently, Yuan et al.~\cite{yuan2022compositional} propose an end-to-end training method called compositional training for DAM. The idea is to solve a compositional objective $L_{\text{AUC}}(\w - \alpha\nabla L_{\text{CE}}(\w))$, where $L_{\text{AUC}}$ denotes an AUC loss and $L_{\text{CE}}$ denotes a standard cross-entropy loss. It is shown that the compositional training method for DAM yields much better feature representations than optimizing either the CE loss or the AUC loss from scratch. It remains an open problem how to understand this method theoretically. 
Another direction is to consider self-supervised pre-training methods on large-scale unlabeled medical datasets. This approach was recently explored in~\cite{DBLP:journals/corr/abs-2010-05352,DBLP:journals/corr/abs-2010-00747,azizi2021big}. The success on downstream tasks of using DAM has also been demonstrated for detecting COVID-19 based on X-ray images~\cite{he2021performance}. Nevertheless, we could consider pre-training on much larger medical datasets than those used in existing studies and demonstrate the performance of DAM on multiple downstream medical image classification tasks.

\vspace*{-0.1in}
\paragraph{\bf Learning fair and interpretable AI models.} 
Building trustworthy AI is important for many domains, e.g., healthcare, in particular medical image classification. Two issues are of foremost importance, namely fairness~\cite{barocas-hardt-narayanan,DBLP:journals/corr/abs-1810-01943,DBLP:journals/corr/abs-2010-04053,DBLP:journals/corr/abs-1908-09635} and interpretability~\cite{chen2018looks,hase2019interpretable,arik2020protoattend}.  Although these issues have received tremendous attention in the literature for medical image classification~\cite{DBLP:journals/corr/abs-2102-06764,DBLP:journals/corr/abs-2003-00827,schutte2021using}, developing fair and interpretable DAM methods remains to be explored. 
Some outstanding questions and  work include 
(i) how to develop scalable in-processing algorithms for optimizing AUC under AUC-based fairness constraints~\cite{borkan2019nuanced,kallus2019fairness}; (ii) how to develop scalable and interpretable DAM methods; (iii) evaluating these fairness-aware and interpretable AUC optimizaiton methods on large-scale medical image datasets. 

\vspace*{-0.1in}
{\paragraph{\bf Out-of-Distribution Robustness} An emerging issue in machine learning that has attracted great attention is how to tackle the distributional shifts, i.e., the distribution of testing data differs from that of training data. While this issue has been investigated for traditional risk minimization, it has been rarely explored for AUC maximization. Given that AUC maximization is more aggressive in pushing positive examples ranked above negative examples~\cite{DBLP:journals/corr/abs-2012-03173}, it might cause more severe performance degradation in the presence of distributional shifts. Different types of distributional shifts have been studied, e.g., domain generalization, subpopulation shift, covariate shift, concept drift, etc.  Accordingly, various benchmark datasets following different distributional shifts have been curated~\cite{good,wilds,hu2020ogb,oodbench}. Many of these datasets use AUROC as the performance measure. It remains an open problem how robust are existing DAM methods in the presence of distributional shifts and how to make them more robust. }

Finally, we would like to point out that the above list of issues is not complete. There must be some other issues related to DAM or in the context of DAM to be addressed in the future. While this issue has been studied for traditional risk optimization, it has been rarely explored for AUC maximization. 

\section{Conclusions}\label{sec:con}
In this paper, we have presented a comprehensive survey of AUC maximization methods in the past twenty years with a focus on recent research and development of stochastic AUC maximization and deep AUC maximization. We have compared different methods from different perspectives, e.g., formulations, per-iteration complexity, sample complexities, optimization error, statistical error, empirical performance, etc. We also discuss remaining and emerging issues in deep AUC maximization, and provide suggestions of topics for future work.

\begin{acks}
We thank the editors and anonymous reviewers for their constructive comments. T. Yang is supported by NSF Grant 2110545, NSF Career Award 1844403, and NSF Grant 1933212. Y. Ying is supported by NSF grants (IIS-1816227, IIS-2008532, IIS-2110546, and DMS-2110836).
\end{acks}

\bibliographystyle{ACM-Reference-Format}
\bibliography{AUC}


\end{document}